\pdfoutput=1
\documentclass[10pt, a4paper]{article}

\usepackage[]{lrec-coling2024} % this is the new style

\usepackage{amsmath}
\usepackage{graphicx}
\usepackage{tabularx}
\usepackage{soul}
\usepackage{booktabs}
\usepackage[misc]{ifsym}
\usepackage{subcaption}
\usepackage{algorithm} 
\usepackage{algorithmic}
\usepackage{xspace}
\usepackage{multirow}
\usepackage{makecell}
\usepackage{pifont}
\usepackage{xcolor}
\usepackage{xstring}
\usepackage{color}
\usepackage{fdsymbol}
\usepackage{fontawesome}

\definecolor{ForestGreen}{RGB}{34,139,34}
\definecolor{BrickRed}{rgb}{.72,0,0}
\definecolor{LakeBlue}{RGB}{0,61,153}

\newcommand{\greencheck}{\textcolor{ForestGreen}{\ding{51}}}
\newcommand{\redcross}{\textcolor{BrickRed}{\ding{55}}}

\usepackage{xcolor}
\usepackage{hyperref}
 \definecolor{darkblue}{rgb}{0, 0, 0.5}
  \hypersetup{colorlinks=true, citecolor=darkblue, linkcolor=darkblue, urlcolor=darkblue}

\newcommand{\ours}{\textsl{AoR}\xspace}
\newcommand{\wholename}{\textsl{AoR} (Aggregation of Reasoning)\xspace}
\newcommand{\turbo}{\texttt{GPT-3.5-Turbo-0301}\xspace}
\newcommand{\turbon}{\texttt{GPT-3.5-Turbo}\xspace}
\newcommand{\gpt}{\texttt{GPT-4-0314}\xspace} % gpt-4-0314
\newcommand{\gptn}{\texttt{GPT-4}\xspace} % gpt-4-0314
\newcommand{\claude}{\texttt{Claude-2}\xspace}

\newcommand{\corres}{\textsuperscript{\fontsize{7pt}{6pt}\selectfont \Letter}}
\newcommand{\fstar}{\textsuperscript{\fontsize{6pt}{6pt}\selectfont \faStarO}}

\title{\textsl{Aggregation of Reasoning}: A Hierarchical Framework for Enhancing Answer Selection in Large Language Models}

\name{Zhangyue Yin\textsuperscript{$\diamondsuit$},
{\bf \large Qiushi Sun\textsuperscript{$\heartsuit$},}
{\bf \large Qipeng Guo\fstar,} 
{\bf \large Zhiyuan Zeng\textsuperscript{$\diamondsuit$}}\\ 
{\bf \large Xiaonan Li\textsuperscript{$\diamondsuit$},} 
{\bf \large Tianxiang Sun\textsuperscript{$\diamondsuit$},}
{\bf \large Cheng Chang\textsuperscript{$\diamondsuit$},}
{\bf \large Qinyuan Cheng\textsuperscript{$\diamondsuit$},}\\
{\bf \large Ding Wang\textsuperscript{$\clubsuit$},}
{\bf \large Xiaofeng Mou\textsuperscript{$\clubsuit$},}
{\bf \large Xipeng Qiu\textsuperscript{$\diamondsuit$}\corres\thanks{\Letter \quad Corresponding author.},} 
{\bf \large Xuanjing Huang\textsuperscript{$\diamondsuit$} }}

\address{
\textsuperscript{$\diamondsuit$}School of Computer Science, Fudan University 
\textsuperscript{$\heartsuit$}National University of Singapore \\
\textsuperscript{\faStarO}Shanghai AI Laboratory 
\textsuperscript{$\clubsuit$}Midea AI Research Center\\
\texttt{\{yinzy21,cengzy23,changc21,chengqy21\}@m.fudan.edu.cn} \quad \texttt{qiushisun@u.nus.edu} \\
\texttt{guoqipeng@pjlab.org.cn} \quad
\texttt{\{ding2.wang, mouxf\}@midea.com} \\
\texttt{\{lixn20,txsun19,xpqiu,xjhuang\}@fudan.edu.cn}
}

\abstract{
Recent advancements in Chain-of-Thought prompting have facilitated significant breakthroughs for Large Language Models (LLMs) in complex reasoning tasks.
Current research enhances the reasoning performance of LLMs by sampling multiple reasoning chains and ensembling based on the answer frequency. 
However, this approach fails in scenarios where the correct answers are in the minority.
We identify this as a primary factor constraining the reasoning capabilities of LLMs, a limitation that cannot be resolved solely based on the predicted answers.
To address this shortcoming, we introduce a hierarchical reasoning aggregation framework \wholename, which selects answers based on the evaluation of reasoning chains.
Additionally, \ours incorporates dynamic sampling, adjusting the number of reasoning chains in accordance with the complexity of the task.
Experimental results on a series of complex reasoning tasks show that \ours outperforms prominent ensemble methods. 
Further analysis reveals that \ours not only adapts various LLMs but also achieves a superior performance ceiling when compared to current methods. 
\\ \newline }

\begin{document}

\maketitleabstract

\section{Introduction}
\label{sec:introduction}

Large Language Models (LLMs) have driven remarkable advancements across various Natural Language Processing (NLP) tasks~\citep{openai2023gpt4, chowdhery2022palm, touvron2023llama, touvron2023llama2, huang2022large, zhao2023survey}.
Nonetheless, there remains a discernible gap between the performance of these models and human-level expertise in reasoning tasks~\citep{cobbe2021gsm8k,valmeekam2022large}, which cannot be bridged merely by increasing the model's scale~\citep{rae2021scaling}.
In this context, the advent of Chain-of-Thought (CoT) prompting~\citep{wei2022chain} technique heralds a stride towards mitigating this disparity.
Rather than employing ``answer-only'' prompts, CoT drives LLMs to generate a series of intermediate steps that lead to the final answer.
By decoupling the problem-solving process, CoT not only simplifies the complexity of each step but also offers a novel perspective to addressing complex reasoning tasks.

Beyond the inherent limitations of LLMs~\citep{yin2023selfaware}, \citet{wang2023sc} observe that the CoT exhibits randomness when utilizing a single reasoning chain.
As a remedy, they propose modulating the sampling temperature to collect a diverse set of reasoning chains, and then select the most consistent answer as the final prediction. 
This ensemble approach based on majority-voting has not only elevated the reasoning capability of LLMs but has also emerged as the predominant paradigm for LLMs in reasoning tasks~\citep{chu2023survey, yu2023better}.

\begin{figure}[t]
  \centering
  \includegraphics[width=\linewidth]{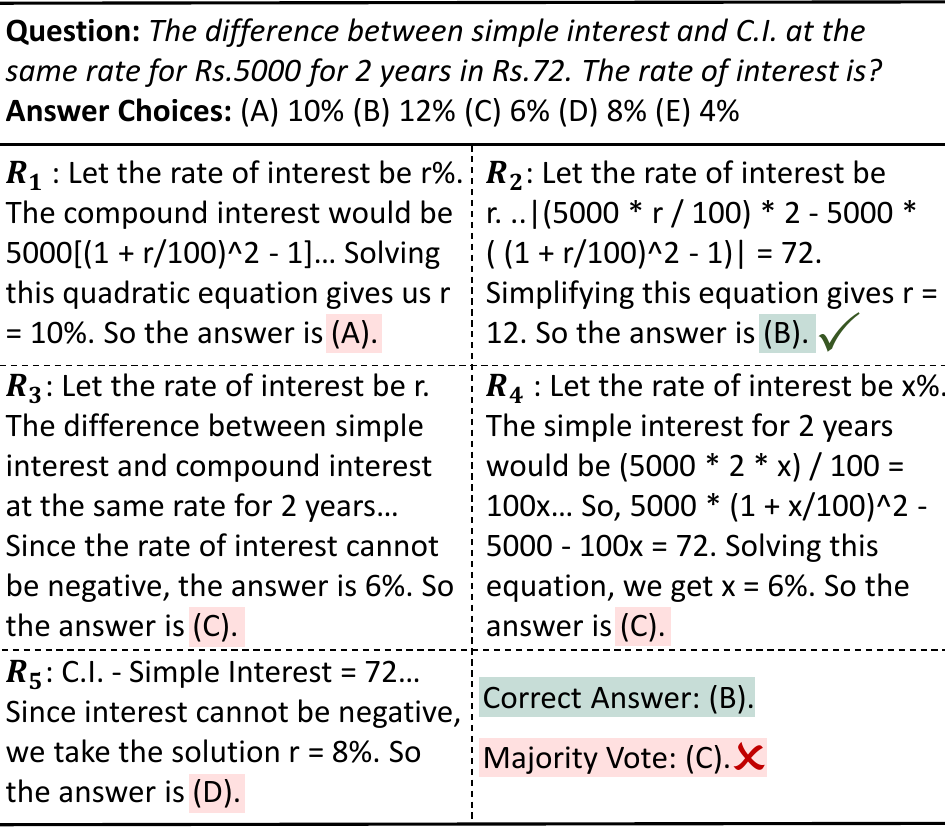}
  \vspace{-1em}
  \caption{An illustrative example from AQuA~\citep{ling2017aqua}, 
  with 5 reasoning chains generated through temperature sampling. 
  Although LLM is able to generate the correct answer, majority voting ultimately selects an incorrect answer due to the abundance of incorrect answers.}
  \label{fig:intro}
  \vspace{-1em}
\end{figure}

\begin{figure*}[thp]
  \centering
  \includegraphics[width=0.95\linewidth]{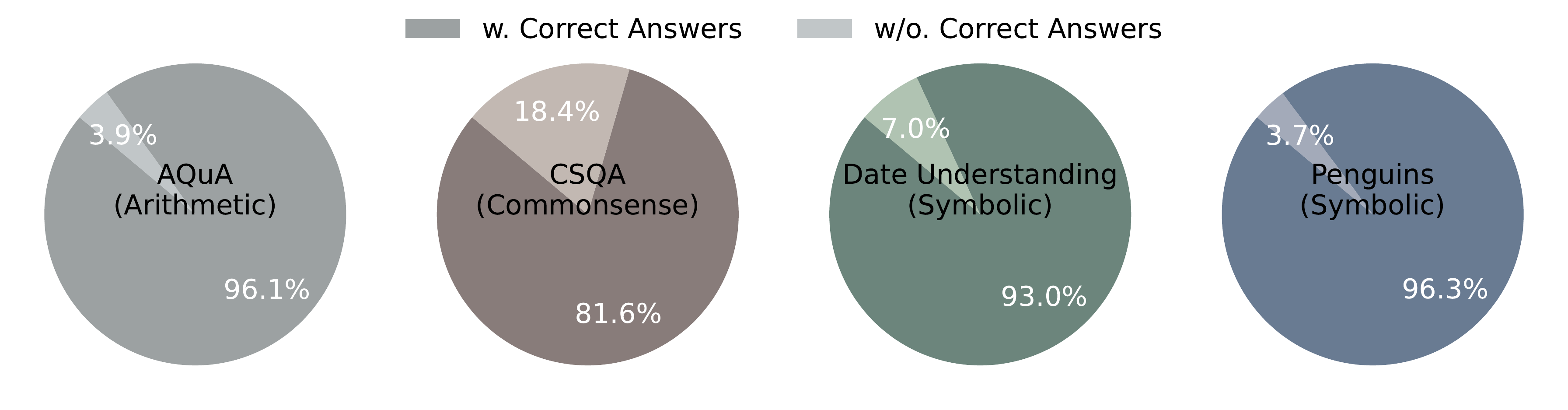}
  \vspace{-1em}
  \caption{Proportion of samples that correct answers appearing in LLMs' generations among those where majority voting results in an incorrect outcome across various reasoning tasks.}
  \vspace{-0.5em}
  \label{fig:intro-data}
\end{figure*}

However, when confronted with more complex questions, LLMs often waver among multiple answers. 
A dilemma arises when the incorrect answers outnumber the correct ones.
Even if the LLM is capable of generating the right answer, the majority voting mechanism remains susceptible to skewing the final prediction towards an erroneous one.
Figure~\ref{fig:intro} showcases an illustrative example from the AQuA dataset~\citep{ling2017aqua}. 
Among the five sampled reasoning chains, four candidate answers: (A), (B), (C), and (D) are generated.
While the LLM is capable of generating the correct answer (B) in \(\mathcal{R}_{2}\) , the overwhelming presence of erroneous candidates eventually led to the selection of the incorrect answer (C).

To explore this phenomenon, we conduct a pilot analysis on samples spanning various reasoning tasks, where the majority voting results in incorrect predictions. 
As depicted in Figure~\ref{fig:intro-data}, over 80\% of the samples that LLM has the potential to answer correctly, but majority voting fails. 
Notably, in AQuA~\citep{ling2017aqua} and Penguins~\citep{suzgun2023bbh} datasets, this proportion exceeds 95\%. 
These findings indicate that ensembling reasoning chains, which relies on the frequency of answers, still has significant room for improvement.

Motivated by the observed limitations, we pose the central research question of this work: \textit{``When LLMs are capable of generating the correct answer, how can we mitigate the interference of incorrect answers to accurately select the right one?''} 
In situations polluted by a myriad of erroneous predictions, relying exclusively on the answers themselves provides limited insight for enhanced accuracy.
Consequently, it becomes both essential and promising to focus on the process leading to these answers: the reasoning chains.
Thus, we introduce a hierarchical reasoning aggregation framework \wholename, %AoR (Aggregation of Reasoning), 
designed to harness the LLM's ability to evaluate reasoning processes in order to improve the selection of the final answer.

Specifically, given the constraints of LLM's context window~\citep{liu2023lost} that prevents simultaneous evaluation of all reasoning chains, \ours initiates by aggregating chains based on their respective answers followed by a two-phase evaluation process. 
In the first phase: local-scoring, chains yielding identical answers are evaluated. 
Since the answers are consistent, the evaluation places greater emphasis on the soundness of the reasoning process and the appropriateness of the reasoning steps. 
For the second phase: global-evaluation, the most logically coherent and methodically valid chains from different answer groups are jointly assessed. 
The objective is to identify the reasoning chain that best exhibits coherence and consistency between the reasoning process and its corresponding answer, thereby designating this answer as the final output.

Furthermore, leveraging the scores derived from the global evaluation phase, \ours can estimate the current confidence level of the LLM in its optimal reasoning process and answer.
This allows \ours to dynamically decide whether it is necessary to sample additional reasoning chains. 
Experimental results across various reasoning tasks demonstrate \ours's effectiveness in significantly enhancing the reasoning performance of LLMs. 
Benefited from dynamic sampling, 
which determines the number of sampling and evaluations by distinguishing between easy and challenging samples, 
\ours also effectively curtails the LLM's reasoning overhead,
establishing a balance between performance and computational cost.

The main contributions are listed below:

\begin{itemize}
    \item{We identify that the existing ensemble mechanism, which solely relies on the frequency of answers, is insufficient. This observation underscores the importance of incorporating the reasoning process, leading to the design of our hierarchical reasoning process aggregation framework \ours.}
    \item{Leveraging the evaluation scores of the optimal reasoning chains, \ours integrates the ability to dynamically sample reasoning chains, efficiently minimizing the reasoning overhead.}
    \item{Extensive experimental results demonstrate \ours's superior performance and cost efficiency compared to existing reasoning chain ensemble methods.}
\end{itemize}
\begin{table*}[!ht]
    \centering

    \resizebox{0.95\linewidth}{!}{
    \begin{tabular}{lccccc}
        \toprule
        \textbf{Feature} & \makecell{\textbf{\ours} \\ (our work)} & \makecell{ \textbf{Self-Consistency} \\\citep{wang2023sc} } & 
        \makecell{ \textbf{ComplexSC} \\ \citep{fu2023complexcot} } & \makecell{ \textbf{PHP}\\\citep{zheng2023php} } & 
         \makecell{ \textbf{\textsc{DiVeRSe}} \\\citep{li2022on} }
        \\
         \cmidrule(lr){1-1}  \cmidrule(lr){2-2}  \cmidrule(lr){3-3}  \cmidrule(lr){4-4}  \cmidrule(lr){5-5} \cmidrule(lr){6-6}
          Task Agnostic?    & \greencheck & \greencheck   & \greencheck    & \redcross & \greencheck  \\ 
          Training-Free?  & \greencheck & \greencheck   & \greencheck    & \greencheck   & \redcross  \\ 
          Plug-and-Play?    & \greencheck & \greencheck   & \greencheck    & \redcross   & \redcross  \\ 
          Dynamic Sampling?  & \greencheck & \redcross & \redcross    & \greencheck   & \redcross  \\ 
         \bottomrule
    \end{tabular}
    }
    \caption{A comparison of \ours to other reasoning chains ensemble methods.}
    \vspace{-.5em}
    \label{tab:dataset-comparison}
\end{table*}

\section{Related work}
\label{sec:related-works}

\vspace{-.5em}
\paragraph{Reasoning with Chain-of-Thought.} 
Chain-of-Thought (CoT; \citealp{wei2022chain}) prompting has emerged as a pivotal technique for eliciting reasoning capabilities in LLMs~\citep{zhao2023survey, liang2023holistic}. 
When guided by samples enriched with explicit reasoning steps, LLMs can produce a series of intermediate steps culminating in a multi-step solution~\citep{zhou2023comprehensive}. 
Remarkably, CoT can enhance the performance of LLMs in reasoning tasks without necessitating additional training~\citep{huang2022towards, min2022rethinking}. 
This characteristic has swiftly garnered widespread attention~\citep{qiao2023reasoning, chu2023survey}, with several studies attributing this phenomenon to the emergent capabilities intrinsic to LLMs~\citep{wei2022emergent, kaplan2020scaling}. 
Subsequent research has concentrated on strengthening the consistency between reasoning paths and answers~\citep{chen2022program, gao2022pal}, automating the construction of prompts~\citep{zhang2022autocot, li2023unified, diao2023active}, eliciting external knowledge~\citep{wang2023cok, li2023mot} and progressively refining the reasoning processes~\citep{yao2023tree, besta2023graph, sel2023algorithm, han2023dialcot, liu2023plan}.

\vspace{-.5em}
\paragraph{Ensemble of Multiple Reasoning Chains.} 
\citet{wang2023sc} identify the randomness in the CoT's single-chain sampling process and subsequently propose the Self-Consistency method. 
This approach entails sampling multiple reasoning chains and selecting the most frequently occurring answer as the final output, which lays the foundation for a series of reasoning chain ensemble methods. 
\citet{fu2023complexcot} observe a positive correlation between the complexity of reasoning chains and the accuracy of generated answers. 
Based on this insight, they propose filtering reasoning chains based on their complexity before employing a majority voting mechanism for the answers. 
Furthermore, \citet{li2022on} train a verifier to score each reasoning chain. 
The answer corresponding to the highest-scoring reasoning chain is selected as the final output. 
From a different perspective, \citet{zheng2023php} suggest using previously generated answers as hints to guide LLMs toward producing accurate answers. 
Furthermore, recent advancements have seen the emergence of strategies encouraging interaction among reasoning chains~\citep{yin2023exchange} or transforming LLMs into multiple agents to benefit from diverse cognitive processes~\citep{sun2023cr}.
A comparison of \ours with some representative reasoning chain ensemble methods is presented in Table~\ref{tab:dataset-comparison}. 
Notably, our method is task-agnostic and does not require additional annotation for training. 
This plug-and-play characteristic, coupled with dynamic sampling, ensures the functionality and cost-effectiveness of our method.

\vspace{-.5em}
\paragraph{Evaluation Capability of LLMs.} 
The automated evaluation capability of LLMs has recently become a prominent point of research~\citep{hackl2023gpt4, hada2023large, zhu2023chatgpt}. 
\citet{liu2023calibrating} and \citet{wang2023chatgpt} discover that LLMs have the potential to produce evaluation results consistent with human experts. 
\citet{chiang2023large} and \citet{shen2023large} further underscores the stability and reliability of assessments generated by LLMs. 
\citet{kocmi2023large} and \citet{liu2023geval} conduct a comparative study between LLM-based evaluation methods and existing automated evaluation metrics. 
Their results showcase that evaluations derived from LLMs surpassed all current automated benchmarks, indicating the exceptional evaluation capabilities of LLMs. 
Moreover, the utilization of LLMs for assessment offers several advantages including customizability~\citep{fu2023gptscore}, a diversity of evaluation perspectives~\citep{chen2023exploring}, and training-free~\citep{luo2023chatgpt}. 
Given the remarkable evaluation prowess of LLMs~\citep{chan2023chateval, chiang2023closer, gao2023humanlike}, we integrate this capability into the aggregation of reasoning chains, enabling a more accurate assessment and selection of the reasoning processes and answers.
\section{Preliminary}
\label{sec:preliminary}

\begin{figure*}[t]
  \centering
  \includegraphics[width=\linewidth]{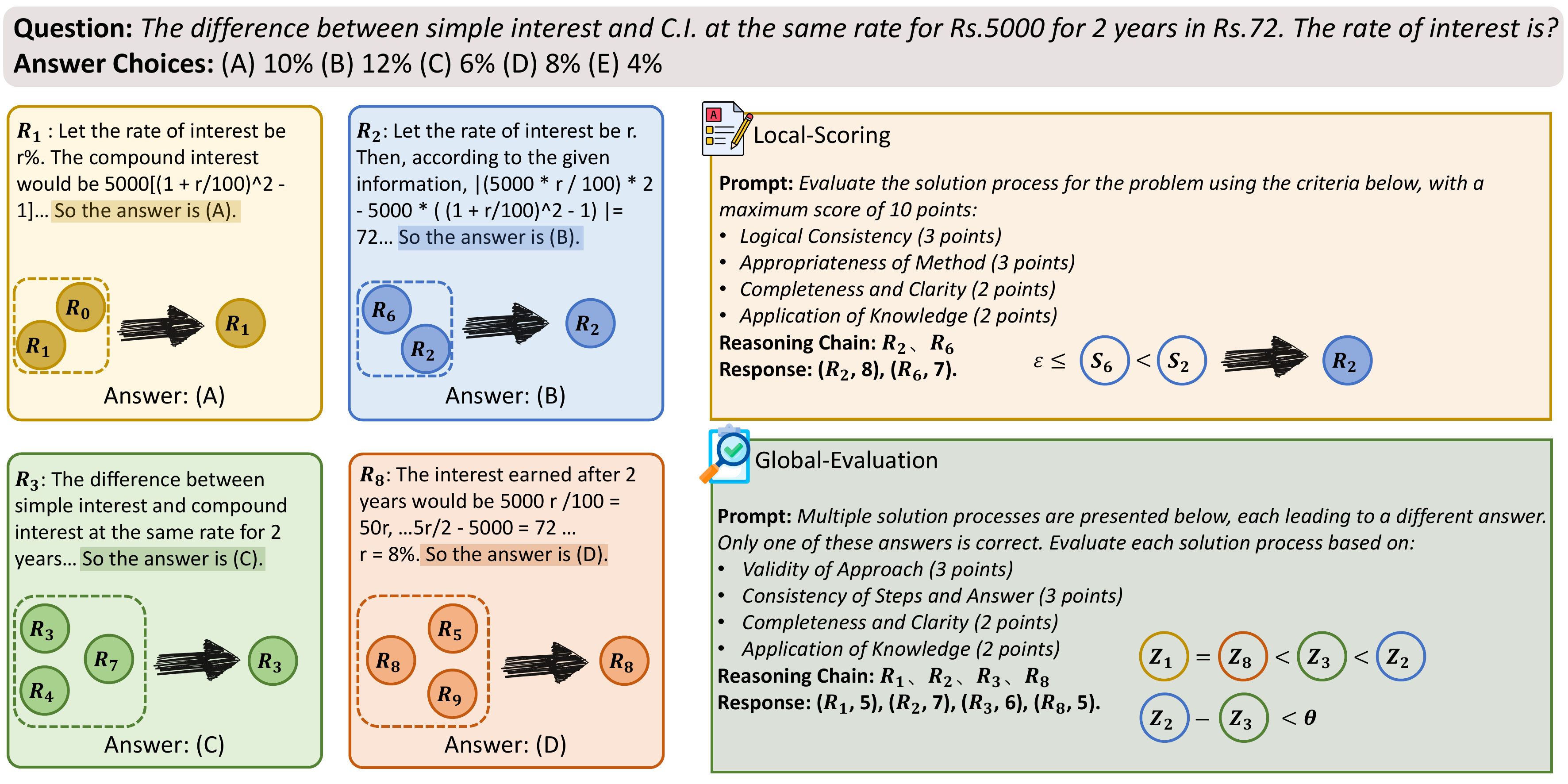}
  \caption{An illustrative example detailing the \ours workflow. Initially, 10 reasoning chains are sampled. During the local-scoring phase, reasoning chains with identical answers are compared, filtering out high-quality chains $\mathcal{R}_1$, $\mathcal{R}_2$, $\mathcal{R}_3$, and $\mathcal{R}_8$ for global evaluation. In the global-evaluation phase, $\mathcal{R}_2$ receives the highest score, but the score margin between $\mathcal{R}_2$ and $\mathcal{R}_3$ fails to surpass the threshold \( \theta \).}
  \vspace{-.5em}
  \label{fig:AoR}
\end{figure*}

In this section, we provide definitions for standard prompting and CoT Prompting. 
Additionally, we detail the voting procedure of Self-Consistency. 
These foundational concepts serve as a groundwork for \ours. 
Considering a scenario where there is a question, denoted as \( \mathcal{Q} \), along with a prompt, denoted as \(\mathcal{T} \), and a LLM, denoted as \(P_{\mathcal{M}} \). 

\vspace{-.5em}
\paragraph{Standard Prompting.}  
Under standard prompting, LLM takes the question $\mathcal{Q}$ and the prompt $\mathcal{T}$ as inputs. 
It then sequentially generates each token of the answer $\mathcal{A}$, aiming to maximize the likelihood at each step.

\begin{align}
    \label{equ:standard_prompting}
    &P(\mathcal{A} ~|~ \mathcal{T, Q}) =
    \prod_{i=1}^{|\mathcal{A}|}
    P_{\mathcal{M}}  
    (a_i ~|~ \mathcal{T, Q}, a_{<i})
\end{align}

\vspace{-.5em}
\paragraph{CoT Prompting.} 
CoT~\citep{wei2022chain} enhances the prompt \( \mathcal{T} \) by integrating the problem-solving process and guiding the LLM to generate a rationale \( \mathcal{R} \) before generating the answer \( \mathcal{A} \). 
We refer to the pair \( (\mathcal{R}, \mathcal{A}) \) as a reasoning chain.

\begin{align}
    \label{equ:cot_prompting}
    &P(\mathcal{R}, \mathcal{A} ~|~ \mathcal{T, Q}) =
    P(\mathcal{A} ~|~ \mathcal{T, Q, R})
    P(\mathcal{R} ~|~ \mathcal{T, Q}), 
    % \\
    % &\label{equ:cot_r}
    % P(\mathcal{R} ~|~ \mathcal{T, Q}) = 
    % \prod_{i=1}^{|\mathcal{R}|}
    % P_{\mathcal{M}}
    % (r_i ~|~ \mathcal{T, Q}, r_{<i})
\end{align}
where \( P(\mathcal{R} ~|~ \mathcal{T, Q}) \) and \( P(\mathcal{A} ~|~ \mathcal{T, Q, R}) \) are defined as follows:

\begin{align}
    \label{equ:cot_definition}
    P(\mathcal{R} ~|~ \mathcal{T, Q}) = 
    \prod_{i=1}^{|\mathcal{R}|}
    P_{\mathcal{M}}
    (r_i ~|~ \mathcal{T, Q}, r_{<i}) \\
    P(\mathcal{A} | \mathcal{T,Q,R}) =
    \prod_{j=1}^{|\mathcal{A}|}
    P_{\mathcal{M}}  
    (a_i~|~\mathcal{T,Q,R}, a_{<j})
\end{align}

\vspace{-.5em}
\paragraph{Self-Consistency.} 
Self-Consistency~\citep{wang2023sc} employs CoT to sample \( n \) reasoning chains: \(\{ (\mathcal{R}_1, \mathcal{A}_1), (\mathcal{R}_2, \mathcal{A}_2), \dots, (\mathcal{R}_n, \mathcal{A}_n) \}\). 
We define the set of answers as \( \{\mathcal{A}\} = \{\mathcal{A}_1, \mathcal{A}_2, \ldots, \mathcal{A}_n\} \). 
The final answer \( \mathcal{A}^{*} \) is determined by selecting the answer that appears most frequently within \( \{\mathcal{A}\} \).

\begin{align}
    \label{equ:self_consistency}
    \mathcal{A}^{*} = \mathop{\arg\max}_{a} |\{(\mathcal{R}_i, \mathcal{A}_i) ~|~ \mathcal{A}_{i} = a\}|
\end{align}
\section{Methodology}
\label{sec:methodology}

\begin{figure*}[t]
  \centering
  \includegraphics[width=\linewidth]{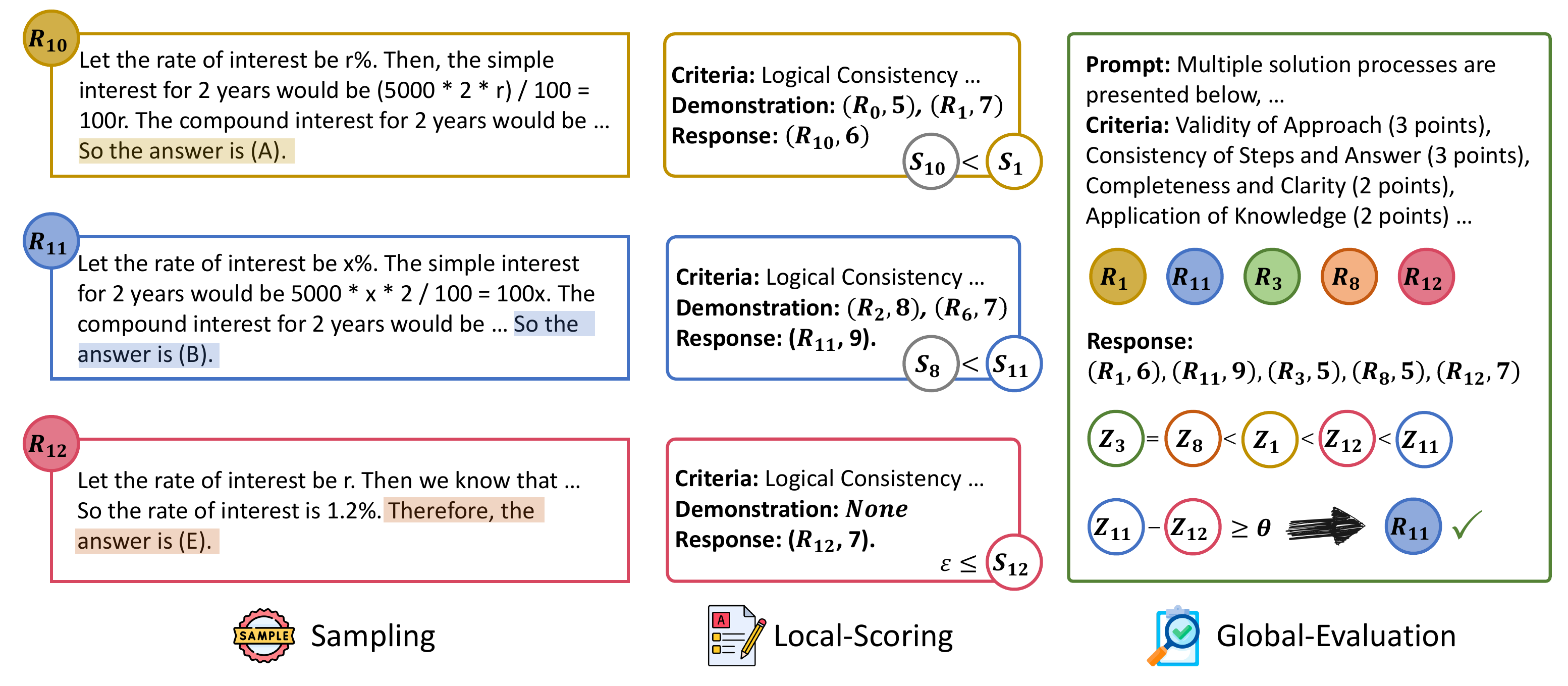} %  \sqs{$\epsilon$}
  \caption{Illustration of the dynamic sampling process, where solid circles represent reasoning chains and hollow circles their respective scores. Due to the minimal score difference between \(\mathcal{R}_2\) and \(\mathcal{R}_3\), three additional chains \(\mathcal{R}_{10}\), \(\mathcal{R}_{11}\), and \(\mathcal{R}_{12}\) are sampled, yielding answers (A), (B), and (E). \(\mathcal{R}_{10}\) and \(\mathcal{R}_{11}\) are compared against chains with matching answers. \(\mathcal{R}_{10}\) fails to outscore \(\mathcal{R}_1\), while \(\mathcal{R}_{11}\) surpasses \(\mathcal{R}_8\), advancing to global evaluation. \(\mathcal{R}_{12}\), introducing a new answer (E), exceeds the threshold \(\epsilon\) and progresses. In the global evaluation, \(\mathcal{R}_{11}\) outperforms others, and with its score difference with \(\mathcal{R}_{12}\) exceeding \(\theta\), thus answer (B) is selected as the final decision.}
  \vspace{-.5em}
  \label{fig:Dynamic-Sampling}
\end{figure*}

\subsection{Overview}
\label{subsec:overview}
The \ours approach to aggregating reasoning primarily unfolds in two stages: local-scoring and global-evaluation. Firstly, we utilize CoT to sample \( n \) reasoning chains, represented as \(\{(\mathcal{R}_1, \mathcal{A}_1), (\mathcal{R}_2, \mathcal{A}_2), \ldots, (\mathcal{R}_n, \mathcal{A}_n)\}\). 
Supposing there are \( m \) unique answers generated, denoted as \( \{a_1, a_2, \ldots, a_m\} \), we categorize them into \( m \) distinct buckets. 
The \( j^{th} \) bucket is defined as \(\{(\mathcal{R}_i, \mathcal{A}_i)~|~\mathcal{A}_i = a_j\}\). 
In the local-scoring phase, we score the reasoning chains \( (\mathcal{R}_i, \mathcal{A}_i) \) within each bucket. 
The top \( k \) chains, based on their scores, are selected as representatives for the bucket. 
In the global-evaluation phase, a representative is selected from each of the buckets for assessment. 
After \( k \) rounds of evaluations, the bucket with the highest average score determines the final output. 
Figure~\ref{fig:AoR} provides an illustrative example. 
Although incorrect answers (C) and (D) are in the majority, the two-phase process of local-scoring and global-evaluation accurately discerns and attributes the highest score to the correct answer (B).

\vspace{-.5em}
\paragraph{Local-Scoring.} Local-scoring focuses on selecting high-quality reasoning chains within a group sharing the same answer. 
While fixing the answer, the evaluation can place a heightened emphasis on the rigor of the rationale logic and the appropriateness of the reasoning steps. 
Let's assume there are \( n_j \) reasoning chains leading to the answer \( a_j \), denoted as \( (\mathcal{R}^{(j)}_1, \mathcal{A}^{(j)}_1), \ldots, (\mathcal{R}^{(j)}_{n_j}, \mathcal{A}^{(j)}_{n_j}) \), collectively forming bucket \( j \).

When these \( n_j \) items are input into the LLM simultaneously, guided by evaluation criteria in the prompt \( \mathcal{T}_1 \), the LLM assigns a score \( \mathcal{S}^{(j)}_i \) to each \( \mathcal{R}^{(j)}_i \). 
Based on a predefined threshold \( \epsilon \), high-quality chains are identified as \( \{(\mathcal{R}^{(j)}_i, \mathcal{A}^{(j)}_i)~|~\mathcal{S}^{(j)}_i \ge \epsilon\} \). 
From this refined set, the top $k$ items are selected as representatives of bucket \( j \), denoted as \( \mathcal{B}^{(j)}_{topk} \). 
If no item satisfies \( \mathcal{S}^{(j)}_i \ge \epsilon \), then \( \mathcal{B}^{(j)}_{topk} \) is an empty set, and items from this bucket will be excluded from the global-evaluation phase.

\vspace{-.5em}
\paragraph{Global-Evaluation.} 
Global-evaluation is tasked with distinguishing and selecting the reasoning chain among different answers,
aiming to pinpoint the one that demonstrates optimal coherence and consistency between the reasoning process and its outcome.
Assuming that we have \( m \) buckets.
A representative is chosen from each bucket, forming a set \(\bigcup_{j=1}^{m} \{b^{(j)} | b^{(j)} \in \mathcal{B}^{(j)}_{topk}\}\).  
When these \( m \) representatives are fed into the LLM concurrently, guided by evaluation criteria encapsulated in prompt \( \mathcal{T}_2 \), the LLM assigns a score \( Z^{(j)} \) to each \( b^{(j)} = (R^{(j)}, A^{(j)}) \).

Representatives from each bucket are sequentially chosen for \( k \) rounds of scoring.
Ultimately, the bucket $j^{*}$ with the highest average score determines the final answer. 
If the number of representatives in a bucket is less than \( k \), previously selected items are resampled to meet the required count.

\begin{align}
    \label{equ:local-scoring}
    &j^{*} = \mathop{\arg\max}_{j} \frac{1}{k} \sum^{k}_{t=1} \mathcal{Z}^{(j)}_t
\end{align}
\vspace{-1em}

It's worth noting that representatives from each bucket are high-quality reasoning chains bearing scores that are identical or closely aligned, each showcasing its unique advantages. 
Consequently, conducting multiple rounds of scoring not only mitigates the randomness in single-round evaluations but also ensures a comprehensive assessment.

\subsection{Dynamic Sampling}
\label{subsec:dynamic-sampling}
Leveraging the scores from the global-evaluation phase, \ours dynamically adjusts the sampling of reasoning chains based on the LLM's confidence in the optimal reasoning chain. This process begins by identifying two key answers: \( \mathcal{A}^{\alpha} \), which has the highest average score \( \bar{\mathcal{Z}}^{\alpha} \), and \( \mathcal{A}^{\beta} \), with the second-highest average score \( \bar{\mathcal{Z}}^{\beta} \). Drawing inspiration from \citet{roth2006margin}, we consider the margin \( \bar{\mathcal{Z}}^{\alpha} - \bar{\mathcal{Z}}^{\beta} \). If this margin exceeds a predefined threshold \( \theta \), it signifies a substantial quality discrepancy between the top two reasoning chains, leading to the selection of \( \mathcal{A}^{\alpha} \) as the final answer and terminating the sampling process.

If \( \bar{\mathcal{Z}}^{\alpha} - \bar{\mathcal{Z}}^{\beta} < \theta \), AoR proceeds to sample an additional \( d \) reasoning chains. These new chains undergo evaluation against established benchmarks to calculate their scores \( \{S_{n+1}, S_{n+2}, \dots, S_{n+d}\} \). These scores determine their influence on the existing answer hierarchy. If \( S_{n+1} \) is either beneath the threshold \( \theta \) or does not surpass the minimum score within the top \( k \) scores of its answer category, the sampled chain \((\mathcal{R}_{n+1}, \mathcal{A}_{n+1}) \) does not affect the overall ranking. Conversely, if a sampled chain introduces a new answer \( \mathcal{A}_{n+1} = a_{m+1} \) satisfied threshold \( \theta \) or significantly alters the score ranking within \( \mathcal{B}_{topk} \), a re-evaluation during the global-evaluation phase is necessitated to recalibrate scores.

Dynamic sampling ceases once the confidence margin between the two leading answers meets or exceeds \( \theta \) or when the total number of sampled chains reaches a predefined maximum \( n_{max} \). As illustrated in Figure~\ref{fig:Dynamic-Sampling},
we present a straightforward instance of dynamic sampling, in which the accuracy of the final decision is enhanced by integrating an additional reasoning chain, confidently pinpointing answer B during the global evaluation phase. This flexible method guarantees a more efficient assessment, reducing unnecessary computational efforts on clear-cut cases and focusing more rigorously on analyzing queries that are complex or have ambiguous interpretations. By adjusting the depth of evaluation according to the estimated complexity of each task, \ours efficiently balances precision in its outcomes with optimal use of computational resources.

\begin{table*}[htp]
\centering
\begin{tabular}{lccccccc}
\toprule
                       & GSM8K               & MultiArith    & SingleEQ      & SVAMP               & AddSub        & AQuA                & Avg            \\ \midrule
CoT                    & 80.0                & 97.7          & 91.9          & 78.1                & 86.6          & 54.7                & 81.50          \\
CoT-PHP                & 84.6                & 98.3          & 93.9          & 83.9                & 86.1          & 65.4                & 85.37          \\
CoT-SC(40)             & 88.9                & 99.3          & 94.5          & 85.9                & 87.6          & 68.7                & 87.48          \\
CoT-CC(40)             & 88.7                & 99.2          & 94.3          & 86.1                & 87.8          & 69.3                & 87.57          \\
CoT-Diverse(40)        & 89.2                & 99.3          & 94.5          & 86.6                & 88.7          & 70.9                & 88.20          \\
CoT-\ours(20, 40)        & \textbf{91.8}       & \textbf{99.8} & \textbf{95.5} & \textbf{89.8}       & \textbf{90.6} & \textbf{75.9}       & \textbf{90.57} \\ \midrule
ComplexCoT             & 82.8                & 97.5          & 92.5          & 81.0                & 85.5          & 57.4                & 82.78          \\
ComplexCoT-PHP         & 85.1                & 98.0          & 92.9          & 83.1                & 85.3          & 60.6                & 84.16          \\
ComplexCoT-SC(40)      & 90.6                & 98.5          & 94.9          & 87.5                & 87.5          & 70.5                & 88.25          \\
ComplexCoT-CC(40)      & 90.5                & 98.3          & 93.3          & 87.2                & 87.5          & 70.0                & 87.80          \\
ComplexCoT-\textsc{DiVeRSe}(40) & 90.8                & 98.7          & 94.3          & 87.8                & 88.2          & 72.9                & 88.78          \\
ComplexCoT-\ours(20, 40) & \underline{\textbf{92.9}} & \textbf{99.5} & \textbf{95.3} & \underline{\textbf{91.0}} & \textbf{89.1} & \underline{\textbf{76.4}} & \textbf{90.70} \\ 
\bottomrule
\end{tabular}
\caption{Comparison of performance (accuracy \(\%\)) between \ours and several strong baselines across six mathematical reasoning datasets. 
The highest accuracy scores are \underline{underlined}. 
Within the same prompt, standout results are highlighted in \textbf{bold}. 
All methods employ a \turbo backbone for a fair comparison. 
Results for ComplexCoT and ComplexCoT-PHP are sourced from \citet{zheng2023php}.
The average performance across datasets is provided for an overall comparison.}
\vspace{-0.25em}
\label{tab:math-reasoning}
\end{table*}
\section{Experiment}
\label{sec:experiment}

\subsection{Experimental Setup}
\label{subsec:experimental-setup}

\vspace{-.5em}
\paragraph{Tasks and Datasets.}
We conduct a comprehensive evaluation of \ours across three types of reasoning tasks. 
(1) \textbf{Mathematical reasoning} incorporates six representative datasets, namely GSM8K~\citeplanguageresource{language-cobbe2021gsm8k}, MultiArith~\citeplanguageresource{language-roy2015multiarith}, SingleEQ~\citeplanguageresource{language-koncel2016mawps}, SVAMP~\citeplanguageresource{language-patel2021svamp}, AddSub~\citeplanguageresource{language-hosseini2014addsub}, and AQuA~\citeplanguageresource{language-ling2017aqua}.
(2) \textbf{Commonsense reasoning} covers StrategyQA~\citeplanguageresource{language-geva2021strategyqa}, CommonsenseQA (CSQA; \citealplanguageresource{language-talmor2019commonsenseqa}), BoolQ~\citeplanguageresource{language-clark2019boolq}, and AI2 Reasoning Challenge (ARC-C)~\citeplanguageresource{language-clark2018think}. 
(3) \textbf{Symbolic reasoning} comprises four datasets derived from BigBench~\citeplanguageresource{language-srivastava2023bb, language-suzgun2023bbh}, including Date Understanding, Penguins in a Table, Colored Objects, and Object Counting.
A comprehensive overview and statistical analysis of the dataset is presented in Appendix~\ref{appendix:dataset-statistics}.

\vspace{-.5em}
\paragraph{Baselines.} 
We compare \ours with several strong baselines detailed in Section~\ref{sec:related-works}. 
These include Chain-of-Thought prompting (CoT; \citealp{wei2022chain}), Complexity-based prompting (ComplexCoT; \citealp{fu2023complexcot}), Self-Consistency (SC; \citealp{wang2023sc}), Complexity-based Consistency (CC; \citealp{fu2023complexcot}), Progressive-Hint Prompting (PHP; \citealp{zheng2023php}) and \textsc{DiVeRSe}~\citep{li2022on}.

In our experiments, we adhere to the settings of Self-Consistency~\citep{wang2023sc} and sampled 40 reasoning chains, denoted as (40). 
Regarding notations, CoT and ComplexCoT represent prompt exemplars with different reasoning complexities, while SC, CC, PHP, and \textsc{DiVeRSe} signify various methods of reasoning chain ensemble. 
For instance, the notation CoT-SC(40) signifies that 40 reasoning chains are generated using CoT prompts, followed by the application of the Self-Consistency method.
For all baselines, we follow their official implementations for fair comparison.

\vspace{-.5em}
\paragraph{Backbone LLMs.}
In the main experiments, we employ \turbo. In the discussion part, we introduce a broader variety of models, including \gpt, \claude, and the open-source model \texttt{LLaMA-2-70B-Chat} and \texttt{Mixtral-8x7B}. We access models from OpenAI and Anthropic using their official APIs, while for \texttt{LLaMA-2-70B-Chat} and \texttt{Mixtral-8x7B}, we utilized model weights and code provided by \citet{touvron2023llama2} and \citet{jiang2024mixtral}.

When sampling various reasoning chains, we configure the temperature setting differently across models to optimize their performance. 
Specifically, for \turbon, \gptn, and \claude, we maintain a temperature of 1. 
For LLaMA, we adhere to its official recommendation by setting the temperature at 0.6, and for \texttt{Mistral}, we opt for a temperature of 0.7 to achieve optimal performance. 

By default, \ours initially samples 20 reasoning chains, implementing a dynamic sampling strategy with an upper limit of \(n_{max} = 40\) and a batch size \(b = 5\), collectively referred to as AoR(20,40). During the local scoring phase, we define a representative count of \(k = 3\) and a scoring threshold of \(\epsilon = 6\). For dynamic sampling, we establish a termination criterion with a threshold of \(\theta = 2\), and with each iteration, we sample an additional 5 reasoning chains. Moreover, we utilize the best and worst reasoning chains within the same answer as evaluation benchmarks to evaluate the newly sampled reasoning chains. 
Our implementation details and hyperparameters analysis are available in Appendix~\ref{appendix:implementation-details} and ~\ref{appendix:ablation-study}.

\begin{figure*}[t]
  \centering
  
  \subcaptionbox{
    Commonsense Reasoning Tasks.\label{fig:commonsense-result}
  }[0.48\linewidth]{
    \includegraphics[width=\linewidth]{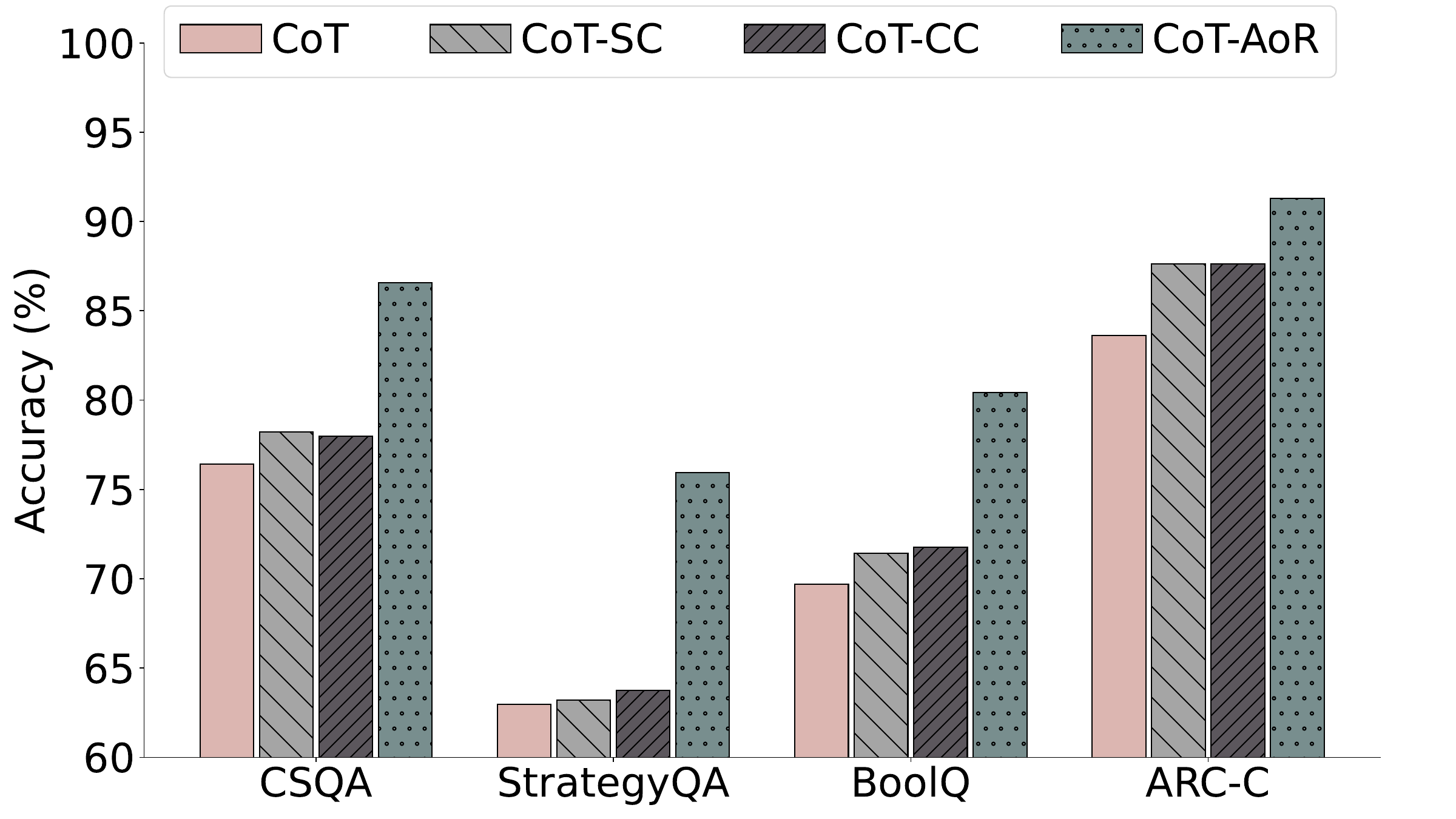}
  }
  \hfill
  \subcaptionbox{
    Symbolic Reasoning Tasks.\label{fig:symbolic-result}
  }[0.48\linewidth]{
    \includegraphics[width=\linewidth]{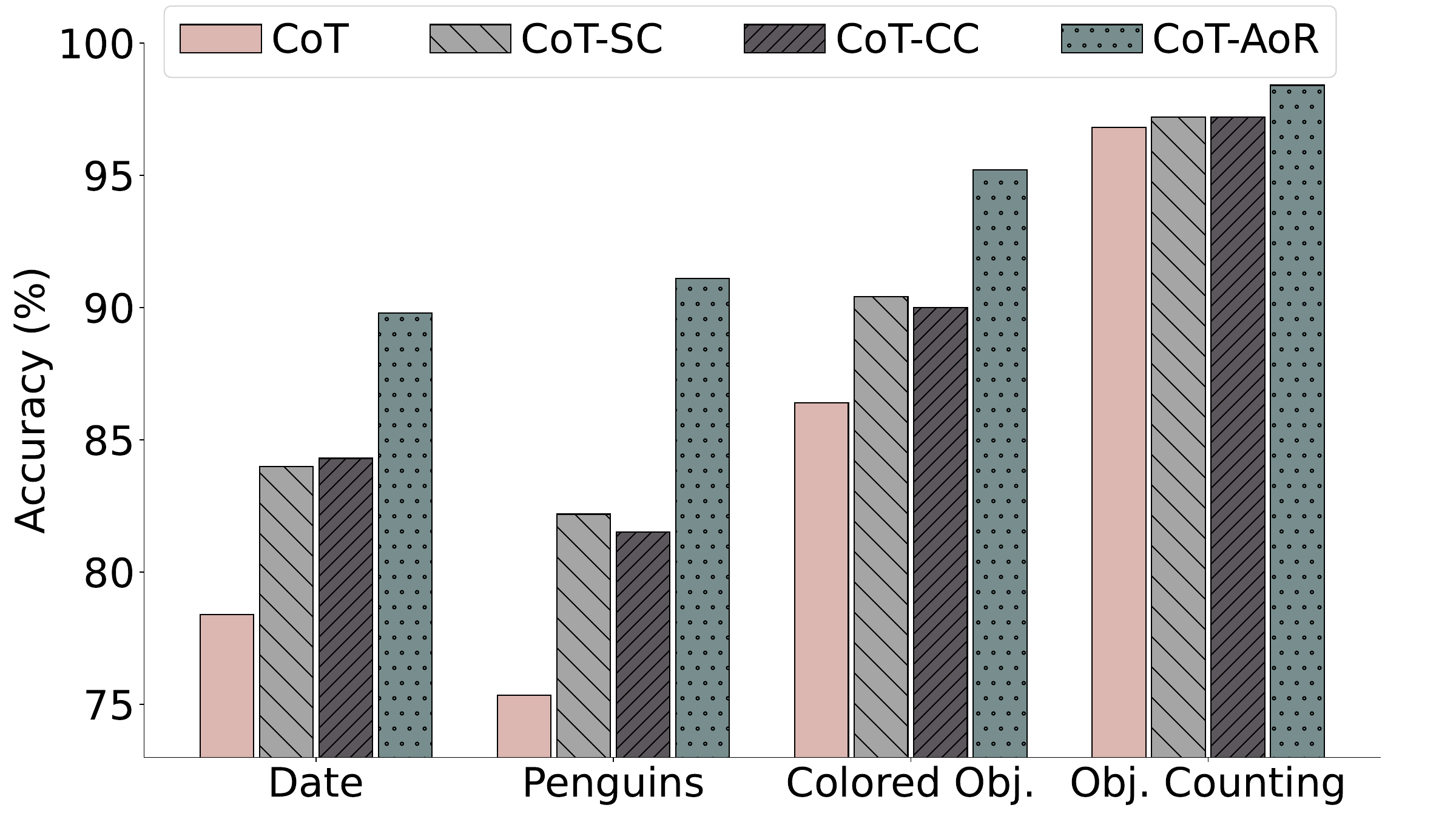}
  }
  \caption{Performance comparison of AoR and various strong baselines on commonsense reasoning and symbolic reasoning tasks.}
  \vspace{-.5em}
  \label{fig:experiment-results}
\end{figure*}

\begin{figure*}[h]
  \centering
  \subcaptionbox{
    AQuA.\label{fig:dynamic-aqua}
  }[0.48\linewidth]{
    \includegraphics[width=\linewidth]{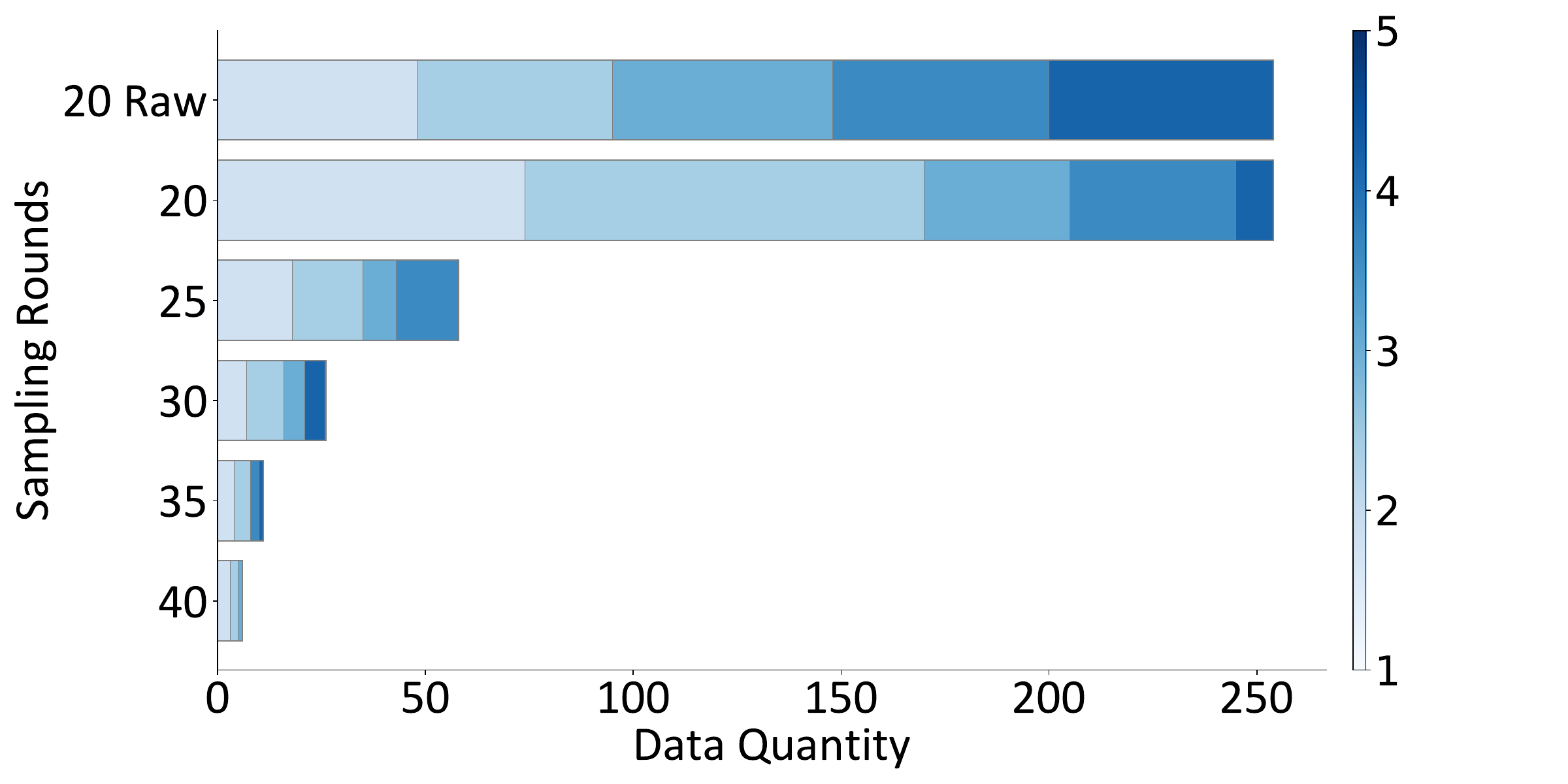}
  }
  \hfill
  \subcaptionbox{
    GSM8K.\label{fig:dynamic-gsm8k}
  }[0.48\linewidth]{
    \includegraphics[width=\linewidth]{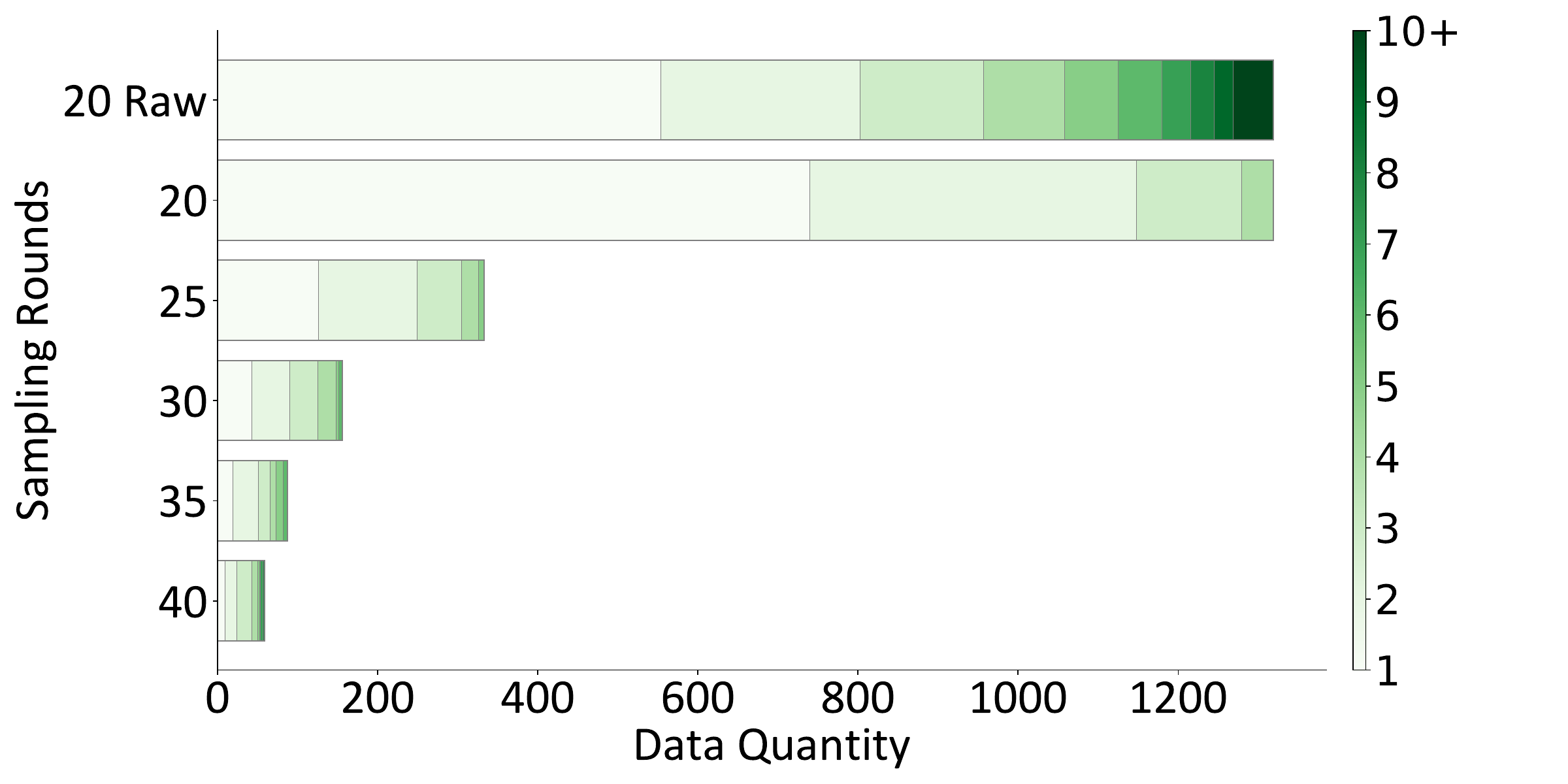}
  }
  \caption{Correlation of sample volume to dynamic sampling iterations in AQuA and GSM8K datasets. The x-axis represents the sample count, while the y-axis indicates the rounds of sampling. Color variations denote the range of answers identified in the global-evaluation phase across different data samples, with ''20 Raw'' indicating the initial distribution of answer counts.}
  \label{fig:dynamic-sampling}
\end{figure*}

\begin{figure*}[t]
  \centering
  \begin{minipage}{.5\linewidth}
    \centering
    \includegraphics[width=\linewidth]{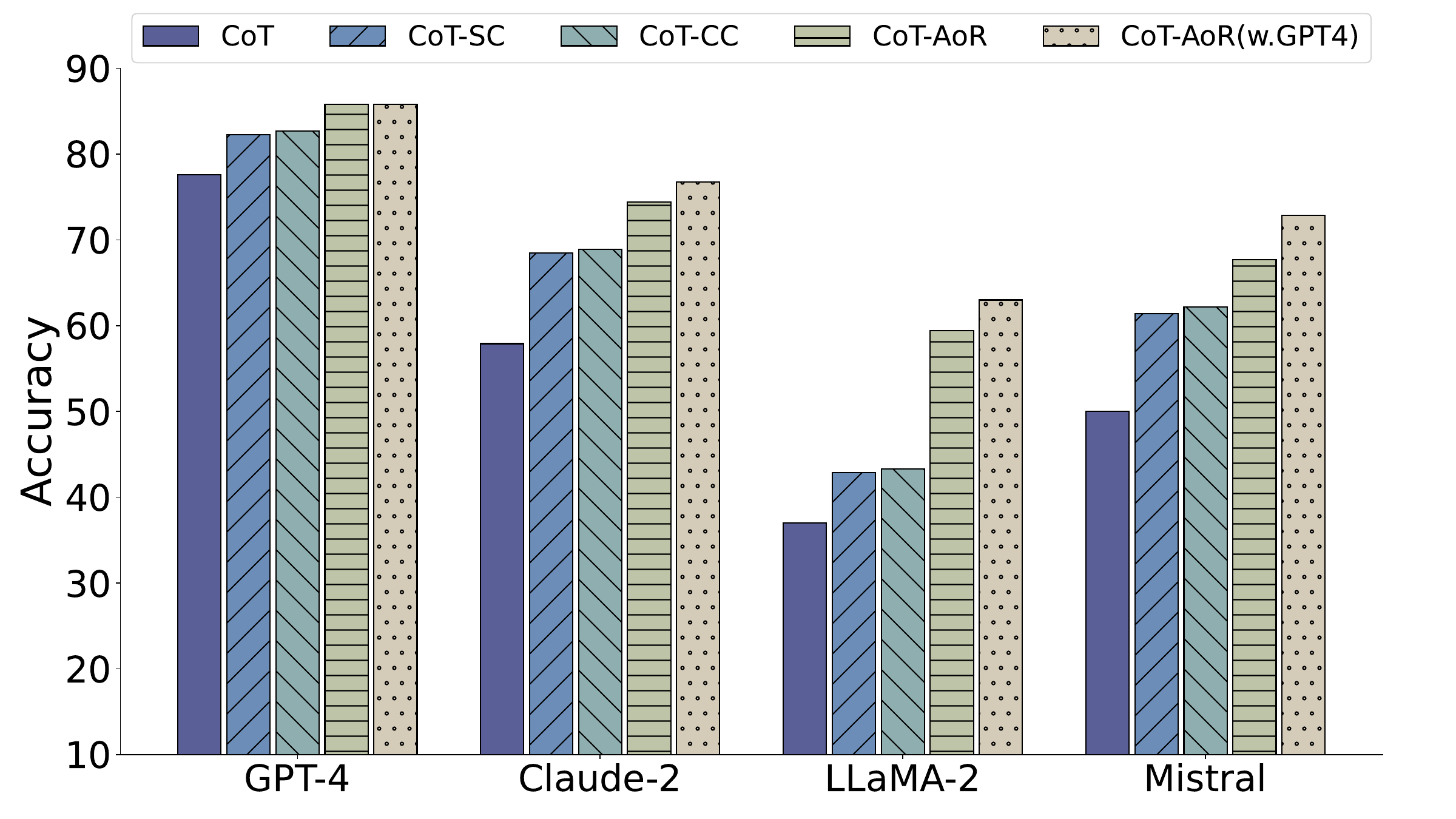}
    \captionsetup{width=.9\linewidth}
    \caption{Performance of \ours using different LLMs for both backbones and evaluators when solving AQuA problems.}
    \label{fig:aor-various}
  \end{minipage}%
  \begin{minipage}{.5\linewidth}
    \centering
    \includegraphics[width=\linewidth]{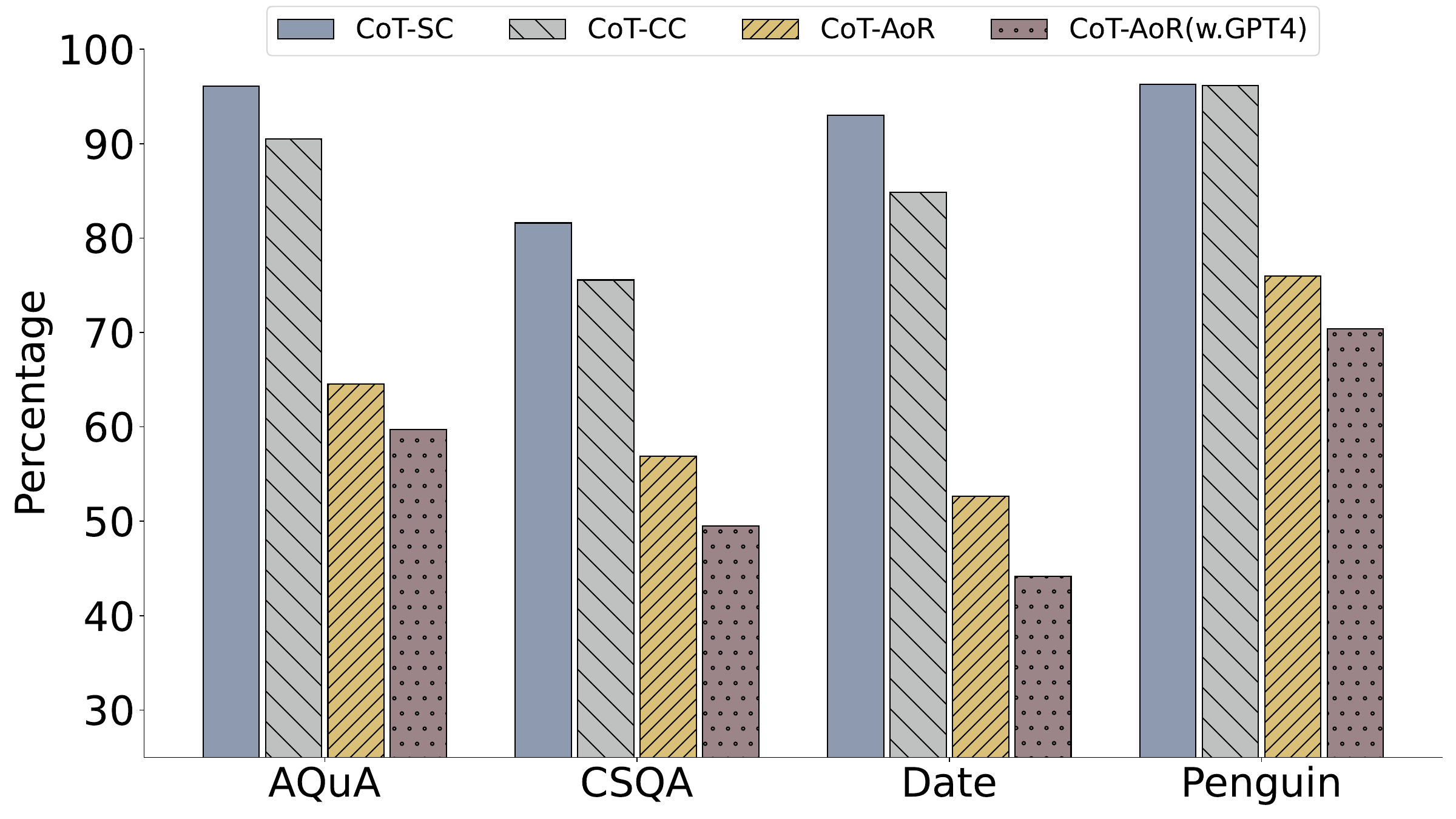}
    \captionsetup{width=.9\linewidth}
    \caption{Proportion of samples lead to incorrect final prediction that contain at least one correct candidate answer.}
    \label{aor-percentage}
  \end{minipage}
\end{figure*}

\subsection{Main Results}
\label{sec:main-results}

\vspace{-.5em}
\paragraph{Mathematical Reasoning.} 
The results for the mathematical reasoning tasks are presented in Table~\ref{tab:math-reasoning}. 
Across six datasets, \ours surpasses all baseline approaches. 
Under the CoT prompt, when compared to the competitive \textsc{DiVeRSe} method, \ours achieves an average performance boost of 2.37\% across six datasets. 
Furthermore, the average performance shows an improvement of 3.09\% compared to the SC method, with a significant increase of 7.2\% on the AQuA dataset. 
When employing ComplexCoT prompt, \ours maintains its competitive advantage. 
It shows average performance enhancements of 2.45\%, 2.90\%, and 1.92\% compared to the SC, CC, and \textsc{DiVeRSe} method. 

\vspace{-.5em}
\paragraph{Commonsense and Symbolic Reasoning.} 
Figures~\ref{fig:commonsense-result} and~\ref{fig:symbolic-result} illustrate the performance of \ours in commonsense reasoning and symbolic reasoning tasks. 
For commonsense reasoning tasks, \ours demonstrate an average performance improvement of 8.45\% and 8.27\% compared to SC and CC methods. 
Notably, on StrategyQA, which emphasizes implicit reasoning strategies, both SC and CC do not significantly outperform the baseline CoT methods. 
In contrast, \ours effectively enhances the LLM's performance on StrategyQA. 
Moreover, \ours consistently achieves significant performance improvements in symbolic reasoning tasks.
When compared to the SC method, there are improvements of 5.8\% and 8.9\% on the Date Understanding and Penguins datasets.

\vspace{-.5em}
\paragraph{Dynamic Sampling.} 
Figures~\ref{fig:dynamic-aqua} and~\ref{fig:dynamic-gsm8k} illustrate the progression of sample counts during the dynamic sampling process on the AQuA and GSM8K datasets. 
The color scheme represents the variance in answer counts within the dataset, transitioning from light to dark shades to illustrate the range from singular to multiple answer occurrences.
The majority of samples conclude satisfactorily after the first round, with only a select group of more complex samples necessitating further reasoning chains.
With each subsequent round of sampling, there's a noticeable decline in the total number of samples, indicating that the newly added reasoning chains contribute to the final answer's determination. 
Compared to the AQuA dataset, which uses options as answers, the open-ended nature of the GSM8K dataset results in a broader distribution of initial answers. 
By observing the distribution of answers in the ``20Raw'' and ``20'' phases, it is evident that in the local-scoring phase, after filtering out a substantial number of low-quality reasoning chains, significantly reduces the number of candidate answers, enabling a more accurate final answer selection in the global evaluation.

\begin{figure*}[t]
  \centering
  \begin{minipage}{.5\linewidth}
    \centering
      \includegraphics[width=\linewidth]{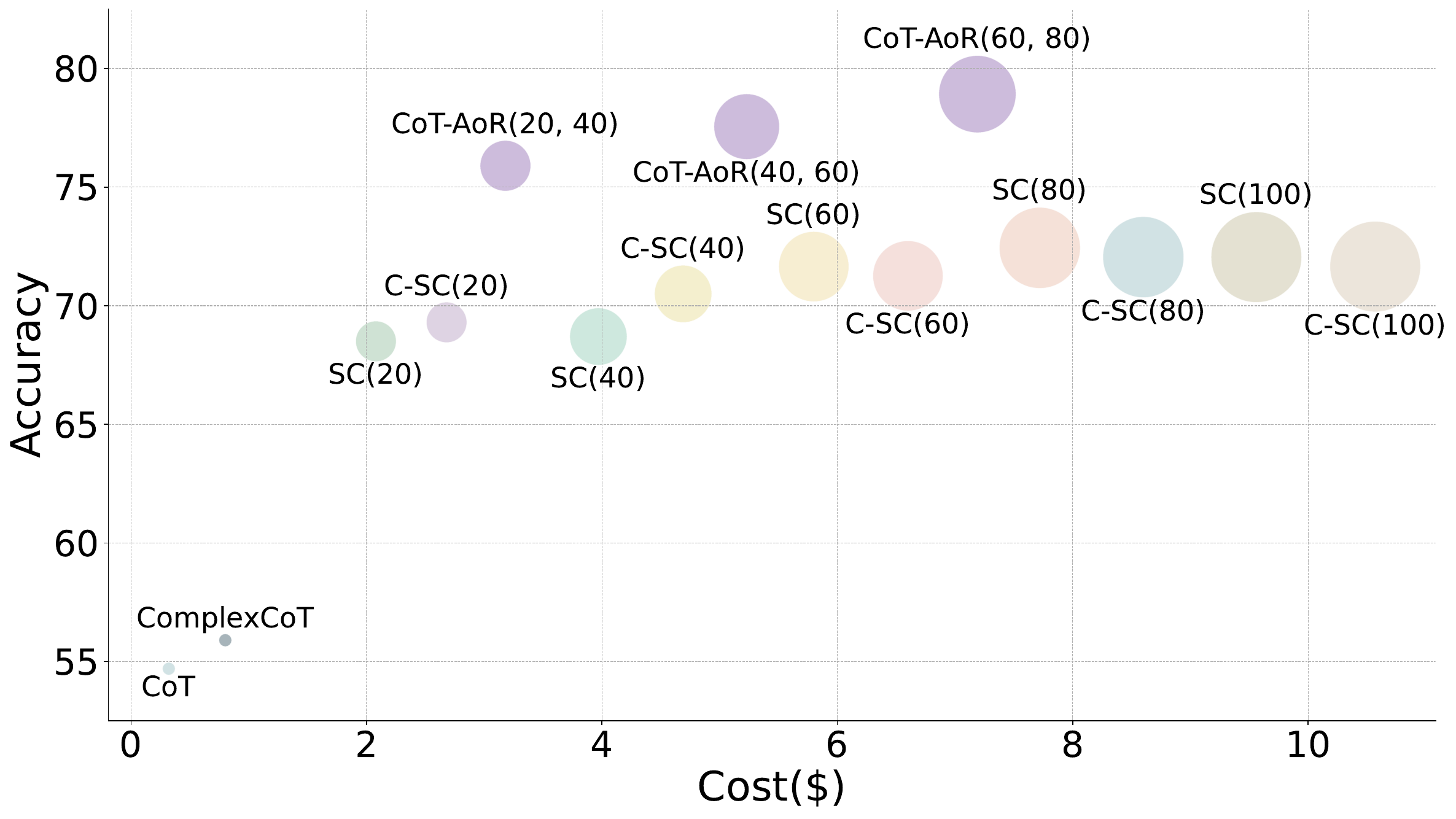}
      \captionsetup{width=.9\linewidth}
      \caption{Analysis of cost and performance. The x-axis represents the cost, the y-axis indicates accuracy, and the size of each point corresponds to the number of reasoning chains.
      For brevity, we use ``SC'' to represent ``CoT-SC'', and ``C-SC'' to denote ``Complexity-SC.''}
      \label{fig:cost-effectiveness}
    \label{fig:aor-various}
  \end{minipage}%
  \begin{minipage}{.5\linewidth}
    \centering
      \includegraphics[width=\linewidth]{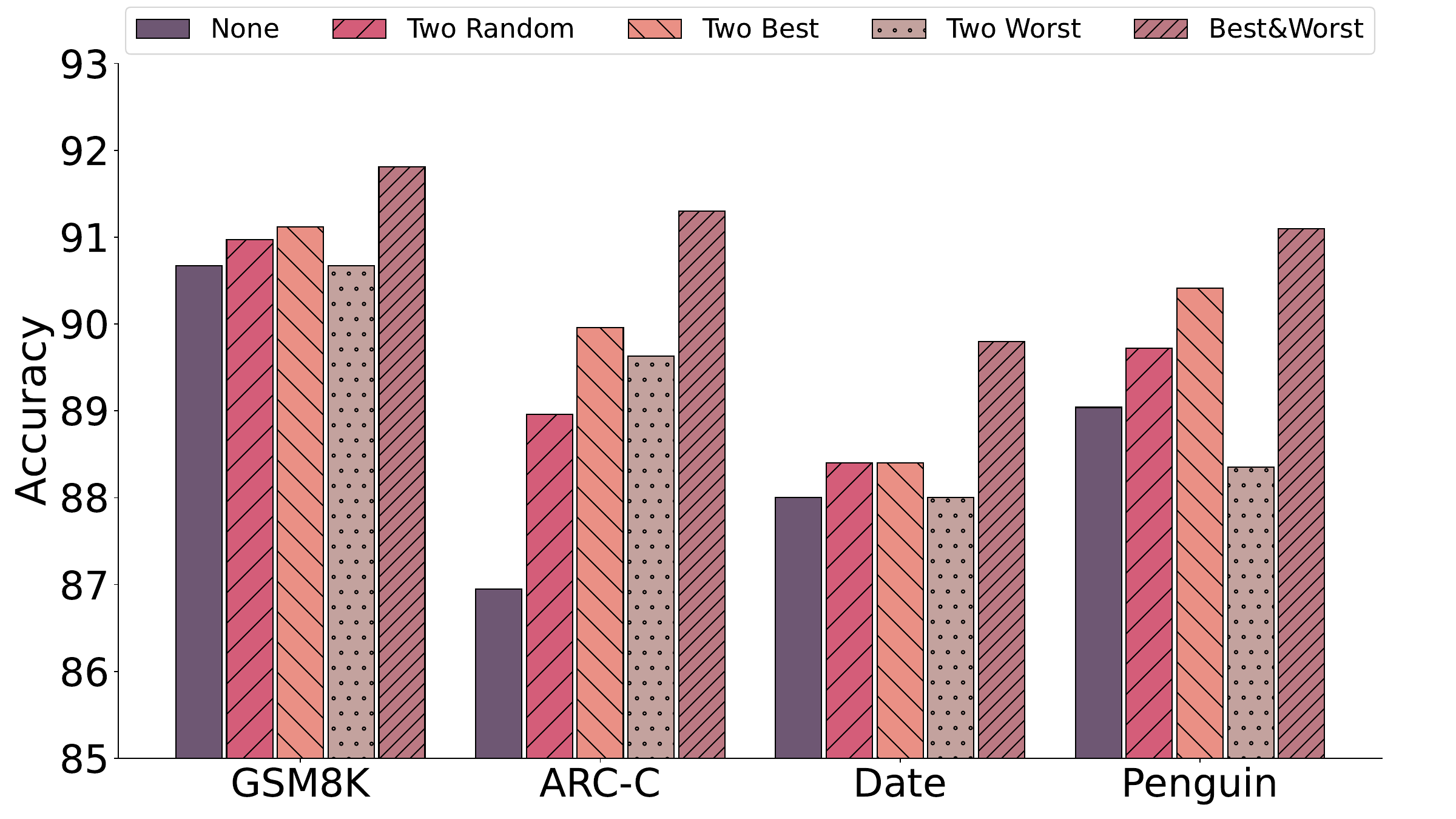}
      \captionsetup{width=.9\linewidth}
    \caption{Evaluation of demonstration selections during the local-scoring phase of dynamic sampling. ``Best'' and ``Worst'' denote the reasoning chains with the highest and lowest scores, respectively, among those yielding identical answers.}
      \label{fig:demonstration-selection}
  \end{minipage}
\end{figure*}

\subsection{Discussion}
\label{subsec:discussion}

In this section, we delve into the advantages of the \ours method, dissecting the reasons behind its performance improvements from four perspectives.

\vspace{-.5em}
\paragraph{\ours on Various LLMs.} 
Figure~\ref{fig:aor-various} depicts the enhanced performance of \ours when applied to four different LLMs.
In comparison with SC and CC, \ours achieves an average improvement of 8.1\% and 7.6\%. 
Our evaluation extends to two prominent open-source models: the dense model \texttt{LLaMA-2-70B-Chat} and the Mixture-of-Experts (MoE) model \texttt{Mixtral-8x7B}.
\ours achieves consistent improvements across various LLM architectures.
Notably, with the \texttt{LLaMA-2} model, the improvement is notably significant, attaining a 16.6\% increase compared to SC. 
Moreover, we conduct an analysis of the evaluation models and obverse that integrating \texttt{GPT-4} into the local-scoring and global-evaluation phases results in performance improvements of 2.4\%, 3.6\% and 5.1\% on the \texttt{Claude-2}, and \texttt{LLaMA-2}, and \texttt{Mistral} models. 
This highlights the potential for a superior evaluation model to enhance the effectiveness of \ours.

\vspace{-.5em}
\paragraph{Analysis of Incorrect Samples.} 
In Section~\ref{sec:introduction}, 
we analyze the erroneous samples from the SC method, revealing that a majority of these samples did not arise from LLM's inability to produce the correct answer. 
Instead, the majority voting mechanism failed to identify the right answer. 
Adopting a similar analytical approach for \ours's incorrect samples, as depicted in Figure~\ref{aor-percentage}, we discover a significant reduction in the proportion of samples where \ours failed to select the correct answer. 
This underscores \ours's efficiency in leveraging the information from reasoning chains to boost the likelihood of selecting the correct answer. 
Moreover, this proportion can be further reduced by employing a more discerning evaluator.

\vspace{-.5em}
\paragraph{Cost and Performance Analysis.} 
A potential concern revolves around the additional overhead introduced by \ours's evaluation and whether dynamic sampling can effectively reduce reasoning costs. 
In Figure~\ref{fig:cost-effectiveness}, we analyze the cost and performance of \ours and SC on the AQuA dataset using \texttt{GPT-3.5}. 
Notably, CoT-\ours(20,40) not only surpasses CoT-SC(40) with a 7.2\% boost in performance but also achieves a significant 20\% reduction in overhead. 
Furthermore, CoT-\ours(20, 40) outperforms even CoT-SC(100), indicating that compared to majority voting, \ours's evaluation of reasoning chains is a more efficient method for answer selection. 
It's noteworthy that the SC method exhibits saturation: there is no significant performance improvement when the number of reasoning chains exceeds 60. 
In contrast, \ours continues to show noticeable performance enhancements at sampling chains of 40 and 60.
This suggests that AoR possesses a superior performance ceiling in comparison to SC approaches, underlining its cost-effectiveness and higher potential for accuracy improvement.

\vspace{-.5em}
\paragraph{Analysis of Evaluation Benchmarks.} 
In the local-scoring phase of dynamic sampling, evaluated reasoning chains are leveraged to score newly added chains. 
Figure~\ref{fig:demonstration-selection} assesses the impact of using no demonstrations versus various demonstration strategies on the final answer across different datasets. 
Strategies include selecting no reasoning chains, two random chains, the two highest, the two lowest, and a combination of the highest and lowest scoring chains for demonstration. 
While this primarily affects dynamically sampled cases, demonstrations consistently enhance model performance across datasets. 
This improvement is likely because demonstrations provide the model with insight into the current score distribution, enabling more informed scoring. 
Notably, employing the highest and lowest scoring chains as demonstrations achieves the best performance, likely because they offer a comprehensive view of the score range, aiding the model in more accurately scoring new chains. 
However, using the two lowest scoring chains as examples tends to bias the model towards lower scores, often preventing these new chains from advancing to the global evaluation phase and thus impairing performance. 
Consequently, we utilize both the best and worst reasoning chains within the same answer as evaluation benchmarks mentioned in Section~\ref{subsec:dynamic-sampling}.

\section{Conclusion}
In this study, we introduce \wholename, a pioneering framework that enhances the ensemble methods for reasoning chains by meticulously evaluating and aggregating the reasoning processes. 
\ours employs a two-phase evaluation approach, assessing reasoning chains from multiple perspectives, 
ensuring the evaluation's validity and comprehensiveness. 
Notably, \ours allows for the dynamic adjustment of the number of reasoning chains according to the complexity of the task, substantially minimizing unnecessary computational overhead. 
Experimental results illustrate that \ours significantly improves the reasoning abilities of LLMs, outperforming several established baselines. 
Furthermore, our in-depth analysis indicates that \ours's adaptability extends across various LLM architectures, with potential for further enhancements through integrating a more robust evaluator and an increased volume of reasoning chains. 
Compared to the existing ensemble methods, \ours not only presents benefits in terms of performance and efficiency but also effectively mitigates the risk of accurate answers being overshadowed by more frequent but incorrect predictions. 
% Through this contribution, we aim to underscore the significance of the reasoning process in the evolution of LLMs and encourage continued exploration into unlocking the full potential reasoning capabilities of these models.
\section*{Ethical Statement}
In developing the \ours framework, our team has prioritized ethical considerations to ensure our work respects privacy and promotes fairness. Specifically, the \ours methodology does not involve the collection or utilization of any personally identifiable information. The design of our experimental prompts has been meticulously crafted to prevent any form of discrimination against individuals or groups, thereby safeguarding against privacy breaches and potential socio-ethical implications. Furthermore, we have conducted an in-depth review of the licenses for all datasets employed in our research, as outlined in Appendix~\ref{appendix:dataset-statistics}.

\section*{Limitations}

\paragraph{Manual Demonstration Construction for Local-Scoring and Global-Evaluation.} Our approach relies on manually crafted demonstrations to guide the model in generating outputs in the desired format for extracting scores. This method's efficacy is contingent on the model's ability to accurately interpret these demonstrations and produce outputs as anticipated. In instances where the model fails to comprehend the demonstrations adequately or deviates from the expected output format, the performance of \ours becomes unstable, potentially hindering the completion of its process. Nonetheless, we are optimistic that the evolution of LLMs will bolster their comprehension~\citep{cheng2024adapting, naveed2024comprehensive} and output formatting capabilities~\citep{liang2024controlled, dekoninck2024controlled}, thereby mitigating this issue over time.

\paragraph{Model Context Window Size Limitations.} The limitations imposed by the model's context window size restrict the number of examples that can be processed simultaneously. At present, models face challenges in handling an extensive array of reasoning chains, necessitating a balance between performance assessment and computational expenditure. While smaller parameter models can navigate through the AoR process, their ability is often limited to evaluating single reasoning chains, thereby escalating the computational demands of AoR. However, we believe this to be a temporary constraint. Recent models like \texttt{Mistral}~\citep{jiang2023mistral} and \texttt{InternLM}~\citep{2023internlm} have demonstrated evaluation capacities comparable to those of \texttt{GPT} with appropriate prompting. Moreover, we are encouraged by recent advancements that have significantly expanded the models' context windows~\citep{xiao2023efficient, liu2024scaling}. As long-context models continue to evolve~\citep{ratner2023parallel,wang2023augmenting}, we anticipate that AoR will be able to conduct evaluations on larger batches of reasoning chains, substantially reducing computational costs and enhancing efficiency.

\section*{Acknowledgements}
This work was supported by the National Natural Science Foundation of China (No. 62236004). We are grateful to the reviewers for their insightful comments and suggestions, which have significantly improved the quality of this manuscript.

\section*{Bibliographical References}
\label{sec:bibliographical-references}

\bibliographystyle{lrec-coling2024-natbib}
\bibliography{lrec-coling2024-example}

\section*{Language Resource References}
\label{sec:language-resource-references}
\bibliographystylelanguageresource{lrec-coling2024-natbib}
\bibliographylanguageresource{languageresource}

\clearpage
\newpage
\appendix

\section{Appendices}
\label{sec:appendix}

\subsection{Dataset Statistics}
\label{appendix:dataset-statistics}

\begin{table*}[t]
\centering
\footnotesize
\begin{tabular}{l|c|c|c|c|c}
\toprule
Dataset & Reasoning Task & Answer Type & \# Prompts & \# Test & License \\
\midrule
\href{https://github.com/openai/grade-school-math}{GSM8K}~\citeplanguageresource{language-cobbe2021gsm8k} & Arithmetic & Number & 8 & 1,319 & MIT License \\
\href{https://github.com/wangxr14/Algebraic-Word-Problem-Solver}{MultiArith}~\citeplanguageresource{language-roy2015multiarith} & Arithmetic & Number & 8 & 600 & Unspecified \\
\href{https://gitlab.cs.washington.edu/ALGES/TACL2015}{SingleEQ}~\citeplanguageresource{language-koncel2016mawps} & Arithmetic & Number & 8 & 508 & Unspecified \\
\href{https://github.com/wangxr14/Algebraic-Word-Problem-Solver}{AddSub}~\citeplanguageresource{language-hosseini2014addsub} & Arithmetic & Number & 8 & 395 & Unspecified \\
\href{https://github.com/arkilpatel/SVAMP}{SVAMP}~\citeplanguageresource{language-patel2021svamp} & Arithmetic & Number & 8 & 1,000 & MIT License \\
\href{https://github.com/deepmind/AQuA}{AQUA}~\citeplanguageresource{language-ling2017aqua} & Arithmetic & Multi-choice & 4 & 254 & Apache-2.0 \\

\href{https://github.com/eladsegal/strategyqa}{StrategyQA}~\citeplanguageresource{language-geva2021strategyqa} & Commonsense & T/F & 6 & 2,290 &  MIT license \\
\href{https://www.tau-nlp.sites.tau.ac.il/commonsenseqa}{CommonsenseQA}~\citealplanguageresource{language-talmor2019commonsenseqa}  & Commonsense & Multi-choice & 7 & 1,221 &  Unspecified \\
\href{https://github.com/google-research-datasets/boolean-questions}{BoolQ}~\citeplanguageresource{language-clark2019boolq}  & Commonsense & T/F & 4 & 3,270 &  CC BY-SA 3.0 \\
\href{https://github.com/allenai/arc-solvers}{ARC-C}~\citeplanguageresource{language-clark2018think}  & Commonsense & Multi-choice & 4 & 299 &  CC BY-SA 4.0 \\

\href{https://github.com/suzgunmirac/BIG-Bench-Hard}{Date Understanding}~\citeplanguageresource{language-suzgun2023bbh} & Symbolic & Multi-choice & 3 & 250 &  MIT license \\
\href{https://github.com/suzgunmirac/BIG-Bench-Hard}{Penguins in a Table}~\citeplanguageresource{language-suzgun2023bbh} & Symbolic & Multi-choice & 3 & 146 &  MIT license \\
\href{https://github.com/suzgunmirac/BIG-Bench-Hard}{Colored Objects}~\citeplanguageresource{language-suzgun2023bbh} & Symbolic & Multi-choice & 3 & 250 &  MIT license \\
\href{https://github.com/suzgunmirac/BIG-Bench-Hard}{Object Counting}~\citeplanguageresource{language-suzgun2023bbh} & Symbolic & Multi-choice & 3 & 250 &  MIT license \\
\bottomrule
\end{tabular}
\caption{Overview of datasets utilized in our experiments. \# Prompts indicates the number of Chain-of-Thought (CoT)~\citep{wei2022chain} prompting exemplars used for few-shot prompting. \# Test denotes the total count of test samples in each dataset.}
\label{tab:dataset-statistic}
\vspace{-1em}
\end{table*}

In our experiment, we meticulously select 14 datasets encompassing mathematical reasoning, commonsense reasoning, and symbolic reasoning domains. 
The specifics and statistical details of each dataset, including the data source, task type, answer type, number of prompt samples, total test samples, and dataset licenses, are comprehensively outlined in Table~\ref{tab:dataset-statistic}.
\subsection{Implementation Details}
\label{appendix:implementation-details}

\paragraph{Prompting Exemplars.}

AoR utilizes the \citet{wei2022chain} and \citet{fu2023complexcot} provided prompt exemplars to sample reasoning chains, with the number of prompt exemplars for each dataset detailed in Table~\ref{tab:dataset-statistic}. In the local scoring phase, given that the answers are identical, our evaluation focuses more on the soundness of the reasoning process and the correctness of the reasoning method. Specifically, we require the LLM to evaluate reasoning chains that share the same answer from four perspectives as follows:

\begin{itemize}
    \item \textbf{Logical Consistency (3 points):} The coherence and soundness of the reasoning are evaluated to ensure logical progression.
    \item \textbf{Appropriateness of Method (3 points):} The suitability of the used method is verified, emphasizing that the approach is not unnecessarily complex.
    \item \textbf{Completeness and Clarity (2 points):} All necessary steps must be clearly shown without omission, ensuring easy follow-through.
    \item \textbf{Application of Knowledge (2 points):} The correct and relevant application of formulas, theorems, or facts is assessed.
\end{itemize}
The global-evaluation phase prioritizes the correctness of the method and the consistency between reasoning steps and the answer, enabling the model to filter out the correct reasoning chain from those with differing answers. Specifically, we require the LLM to evaluate reasoning chains with different answers from the following four perspectives:
\begin{itemize}
    \item \textbf{Validity of Approach (3 points):} The employed method effectively addresses the problem, confirming the appropriateness of the approach.
    \item \textbf{Consistency of Steps and Answer (3 points):} It is ensured that all steps are not only correct but also consistent with the final answer.
    \item \textbf{Completeness and Clarity (2 points):} Essential steps are delineated and presented unambiguously, maintaining clarity throughout.
    \item \textbf{Application of Knowledge (2 points):} The precision and appropriateness in the use of formulas, theorems, or facts are verified.
\end{itemize}
In line with the findings of \citet{gao2023prompt}, providing as much detailed information as possible in the input facilitates the generation of the desired outcome. Thus, additional statistical information, such as the number of reasoning chains within a bucket and the number of candidate answers, is incorporated into the prompt. For the complete prompt, please refer to our \href{https://github.com/yinzhangyue/AoR}{Github} repository.

\begin{figure*}[t]
  \centering
  \begin{minipage}{.5\linewidth}
    \centering
      \includegraphics[width=\linewidth]{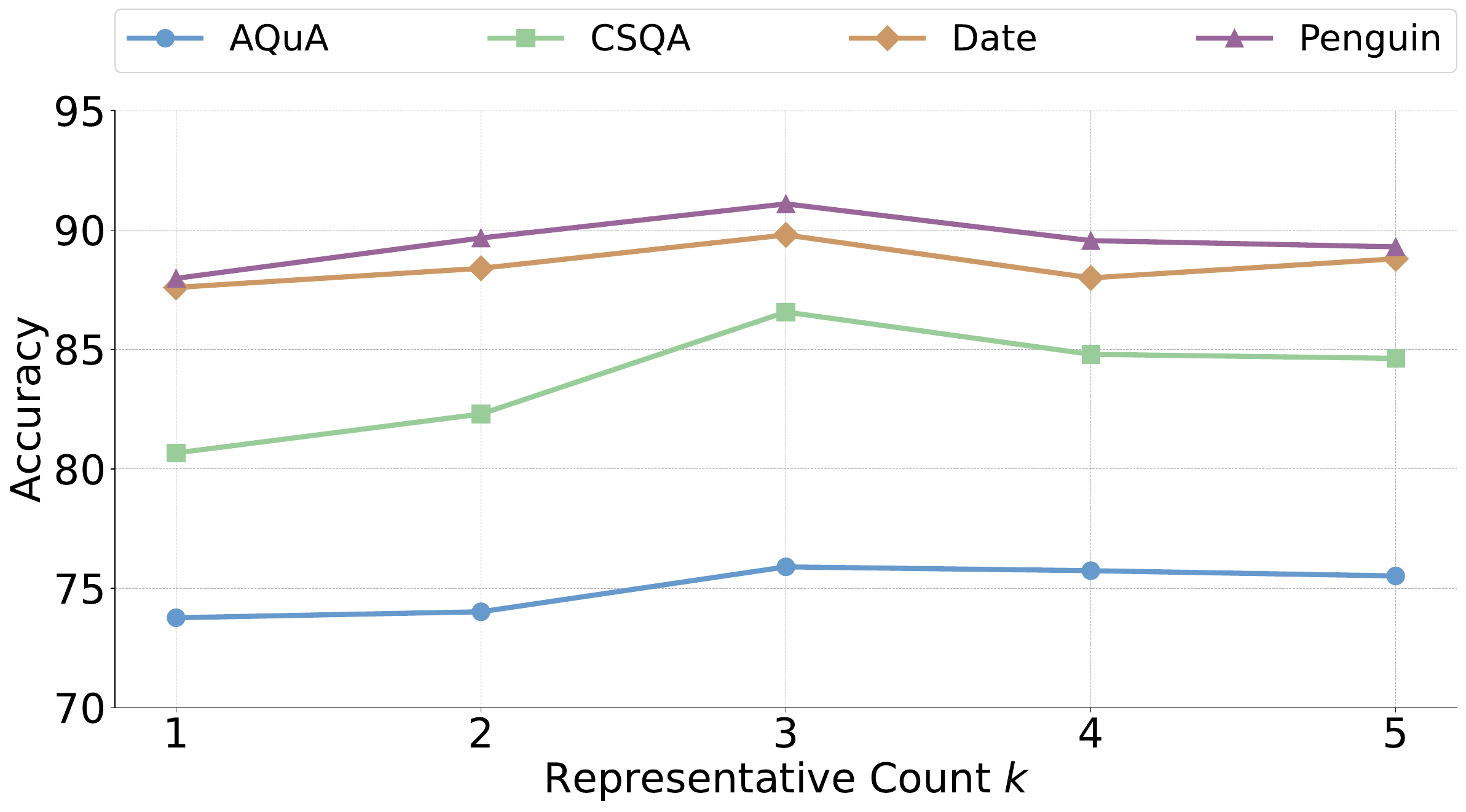}
      \captionsetup{width=.9\linewidth}
    \caption{Ablation on representative count $k$ on various reasoning datasets.}
    \label{fig:representative-count}
  \end{minipage}%
  \begin{minipage}{.5\linewidth}
    \centering
      \includegraphics[width=\linewidth]{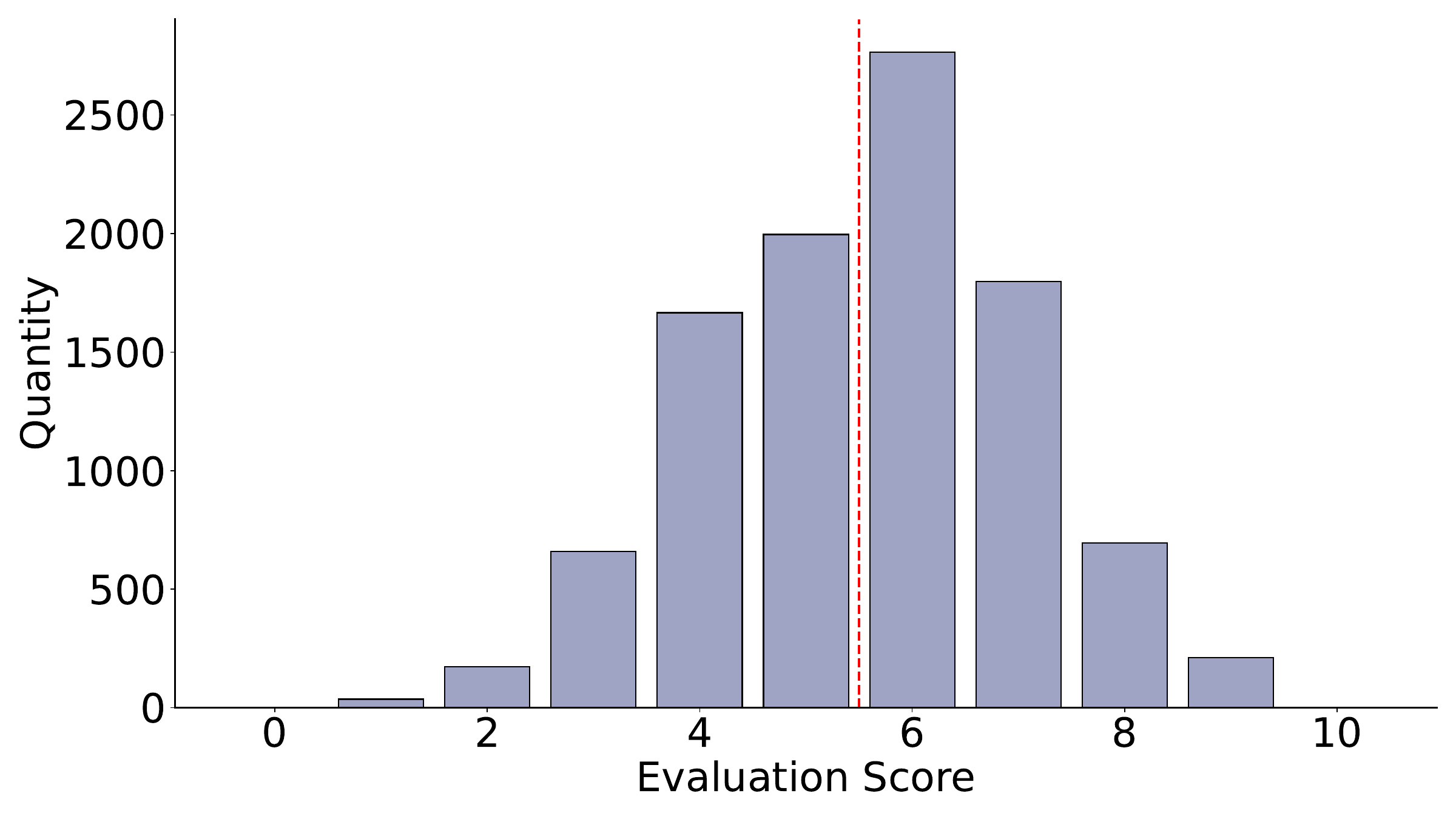}
      \captionsetup{width=.9\linewidth}
    \caption{Distribution of evaluation scores in the local-scoring phase on the GSM8K dataset.}
    \label{fig:score-threshold}
  \end{minipage}
\end{figure*}

\paragraph{Evaluation.}

We employ accuracy as the metric to assess performance across tasks involving mathematical reasoning, commonsense reasoning, and symbolic reasoning. For datasets where the answer is numerical, such as GSM8K, we utilize regular expressions to extract the answer following the phrase ``the answer is'' and conduct a numerical comparison with the provided answer. For datasets where the answers are choices, such as AQuA, we compare the extracted choice with the correct option to verify consistency. In cases where the dataset answers are binary (yes/no), such as StrategyQA, we evaluate whether the extracted result aligns with the provided label. If a reasoning chain fails to correctly extract an answer, it is excluded from further consideration.
Similar to the approach by \citet{xie2023selfevaluation}, we fine-tune task-specific verifiers to assign weights to the sampled reasoning chains to implement the \textsc{DiVeRSe}~\citep{li2022on}.

\paragraph{Computation Cost.}

Computational costs are quantified based on OpenAI's official pricing for the \turbo API, calculated as follows: \(\text{Input Tokens} \times 0.0015 / 1000 + \text{Output Tokens} \times 0.002 / 1000\). 

Our primary experiments, as outlined in Section~\ref{sec:main-results} were conducted from July to September 2023. Discussion in Section~\ref{subsec:discussion} and the Ablation Study in Appendix~\ref{appendix:ablation-study} for both commercial and open-source models were completed between October and December 2023.

Due to rate limits and budget constraints, we set an upper limit on our sample size for each analysis. Consequently, our analysis is based on a maximum of 500 samples per run.

\begin{figure*}[h]
  \centering
  \subcaptionbox{
    GSM8K.\label{fig:termination-threshold-gsm8k}
  }[0.48\linewidth]{
    \includegraphics[width=\linewidth]{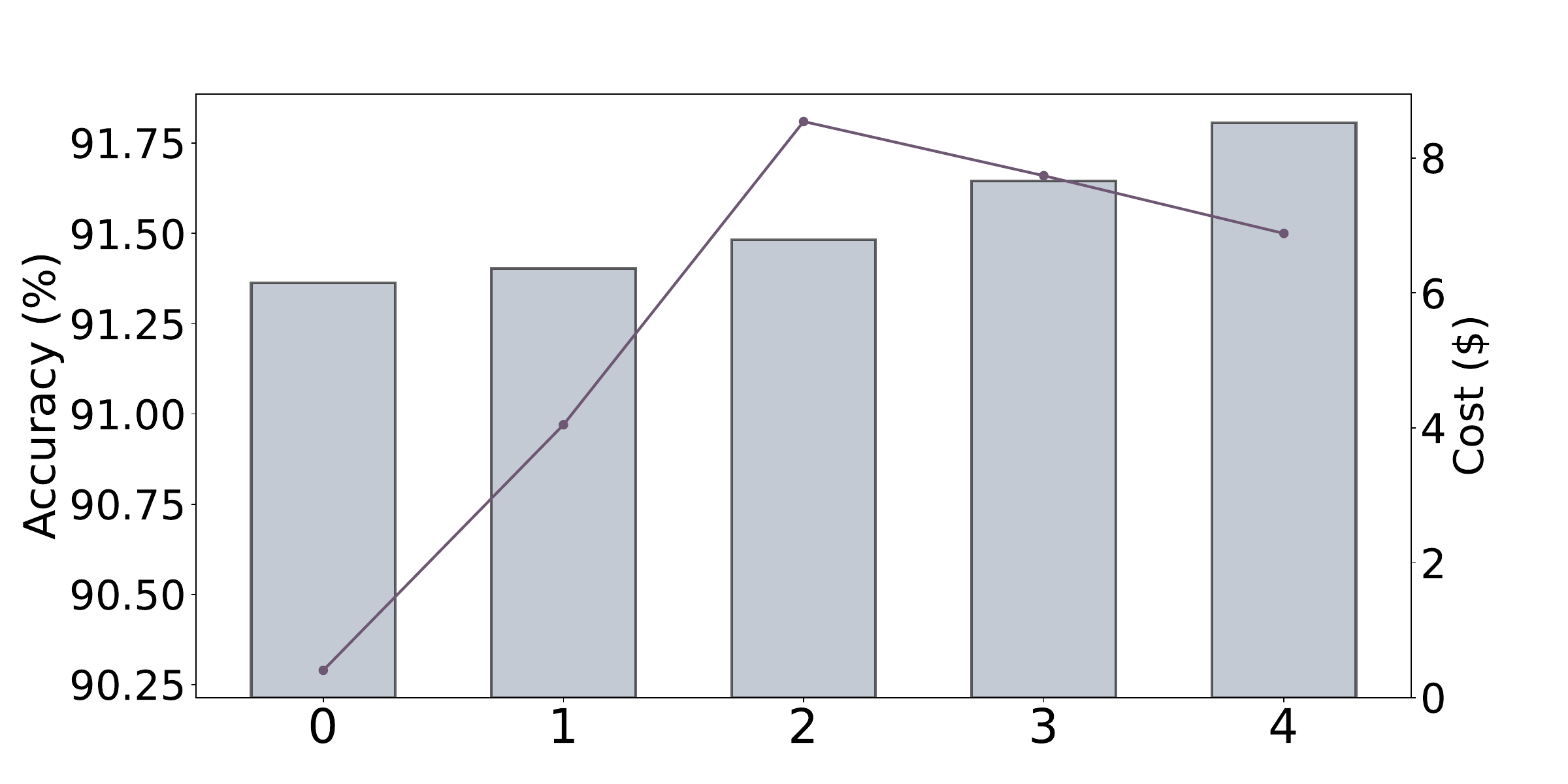}
  }
  \hfill
  \subcaptionbox{
    CSQA.\label{fig:termination-threshold-csqa}
  }[0.48\linewidth]{
    \includegraphics[width=\linewidth]{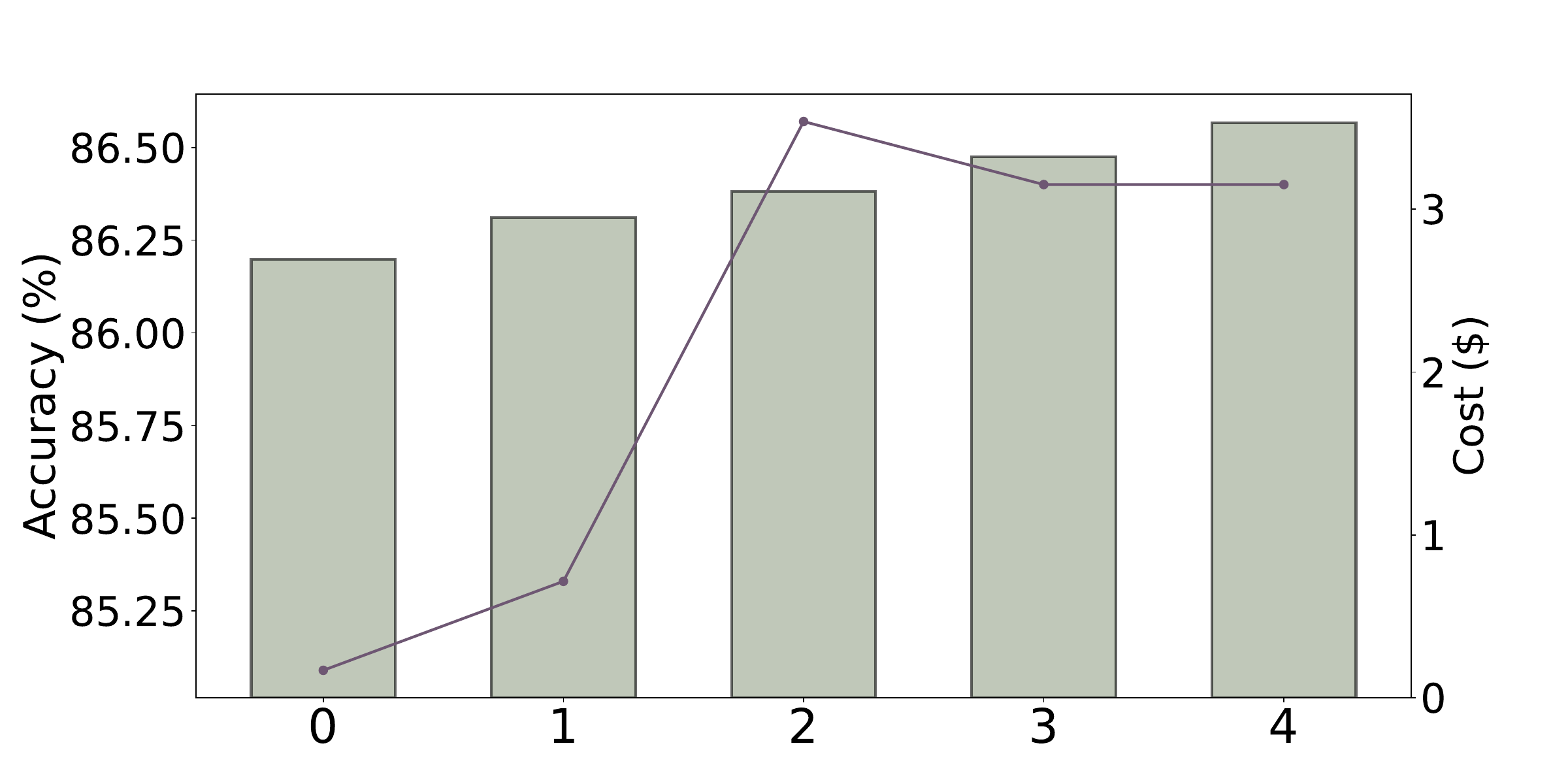}
  }
  \caption{The effect of varying termination thresholds \(\theta\) on accuracy and computational cost for the GSM8K and CSQA datasets. Line graphs illustrate accuracy, while bar graphs depict computational costs.
}
  \vspace{-.5em}
  \label{fig:termination-threshold}
\end{figure*}

\begin{figure*}[h]
  \centering
  \subcaptionbox{
    GSM8K.\label{fig:batch-size-gsm8k}
  }[0.48\linewidth]{
    \includegraphics[width=\linewidth]{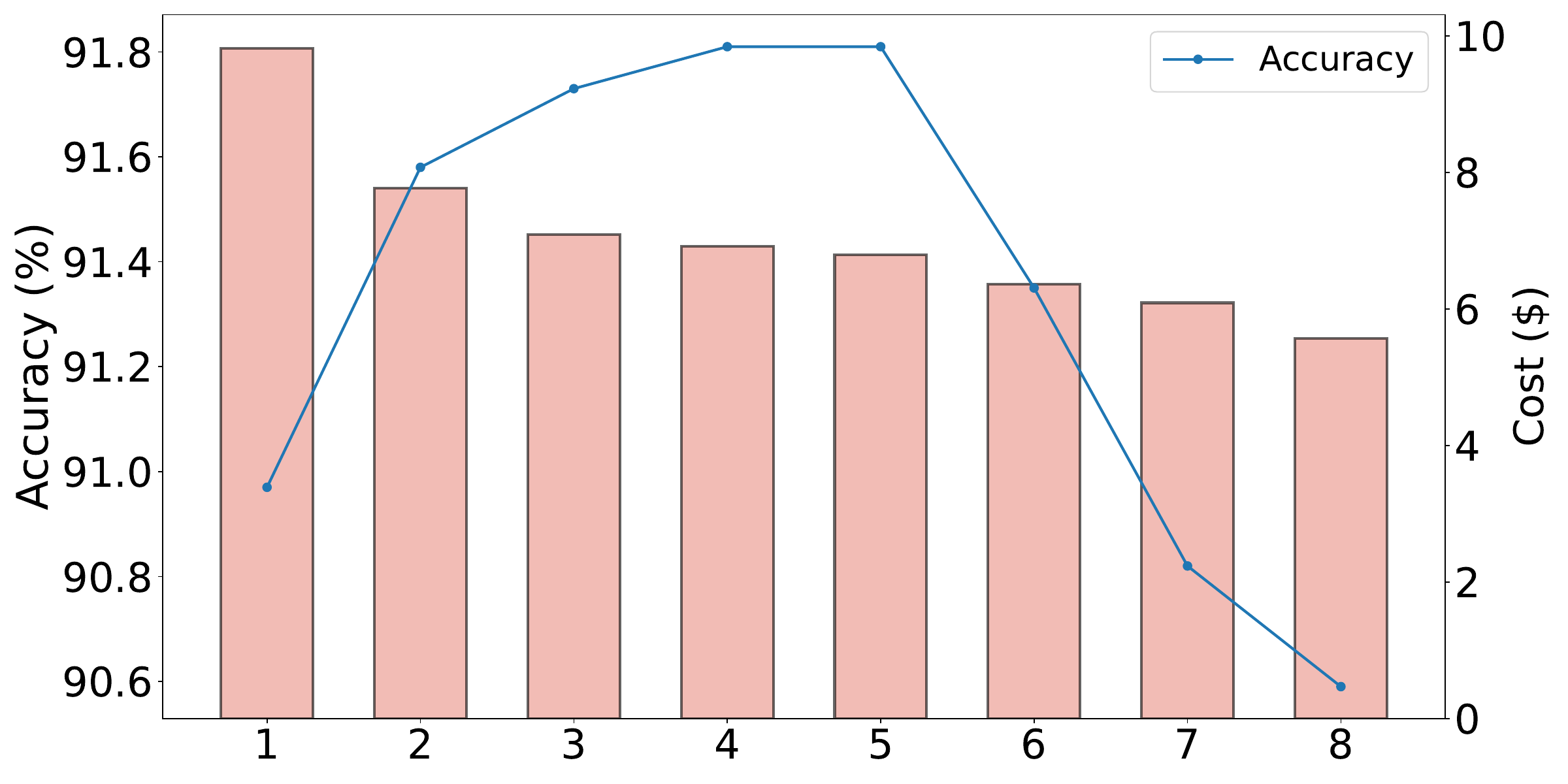}
  }
  \hfill
  \subcaptionbox{
    CSQA.\label{fig:batch-size-csqa}
  }[0.48\linewidth]{
    \includegraphics[width=\linewidth]{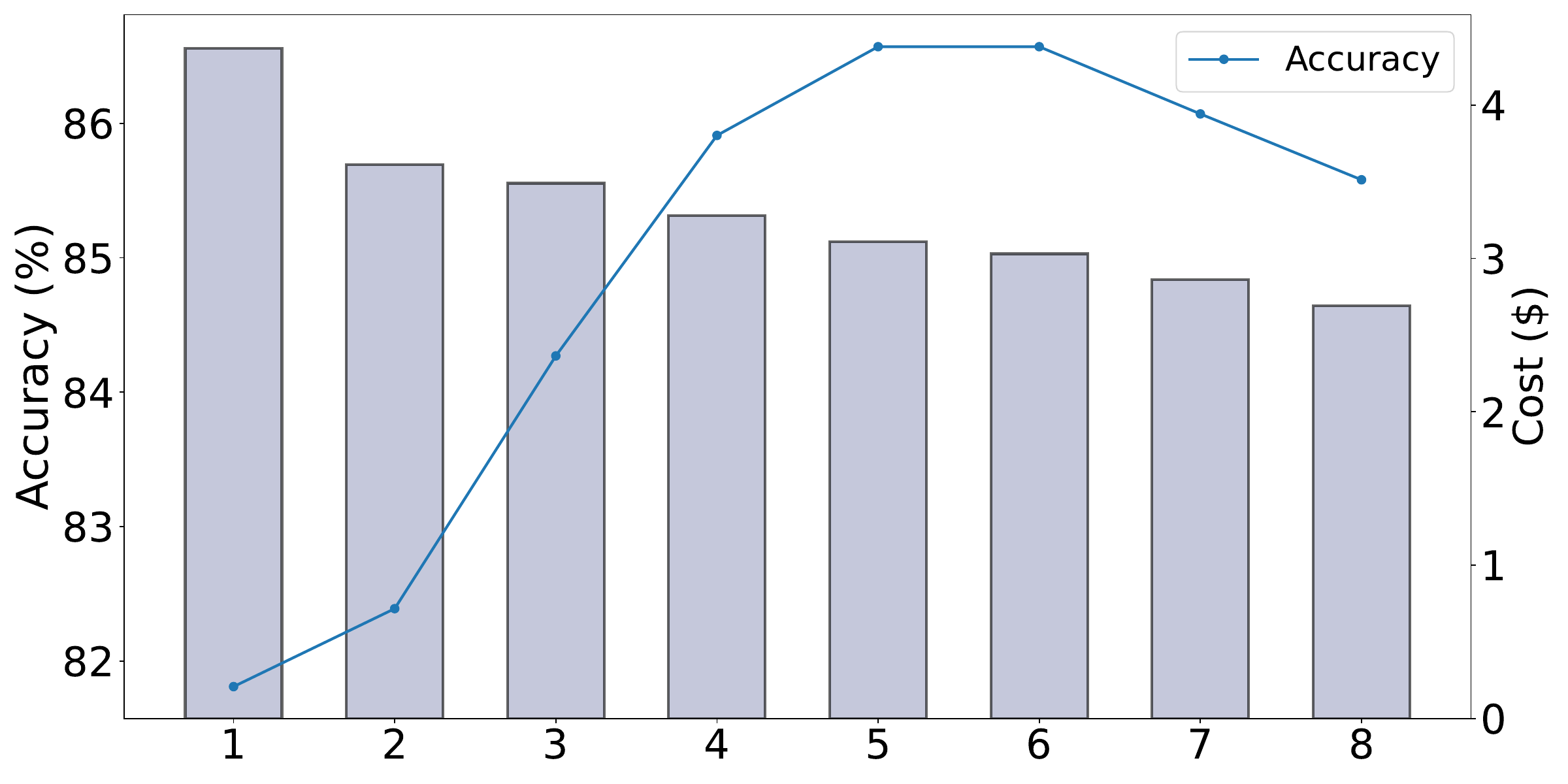}
  }
  \caption{The impact of batch size \(b\) variations on accuracy and computational cost on the GSM8K and CSQA datasets. Line graphs represent accuracy, while bar charts indicate computational costs.
}
  \vspace{-.5em}
  \label{fig:batch-size}
\end{figure*}

\subsection{Ablation Study}
\label{appendix:ablation-study}

To facilitate the intricate reasoning chain aggregation process in AoR, we establish essential hyperparameters during the local scoring and global evaluation phases, such as the representative count \(k\), score threshold \(\epsilon\), termination threshold \(\theta\), and batch size \(b\). Below, we conduct ablation experiments using the \texttt{GPT-3.5} model to examine the impact of each hyperparameter on the overall performance.

\vspace{-.5em}
\paragraph{Analysis of Representative Count \(k\).} 
We analyze four datasets to investigate how the representative count $k$ impacts accuracy, as shown in Figure~\ref{fig:representative-count}. 
When $k=1$, only the highest-scoring reasoning chain from each bucket is evaluated, which puts rigorous demands on the scoring model and can result in fluctuating outcomes. 
Selecting more representatives from each bucket enhances the comprehensiveness and stability of the evaluation, 
harmonizing the quality and diversity of reasoning chains.
However, our findings suggest that increasing the number of representatives when $k>3$ does not lead to significant performance gains but does incur additional computational overhead.
As a result, we chose $k=3$ as it strikes an optimal balance between performance and computational cost.

\vspace{-.5em}
\paragraph{Analysis of Score Threshold \(\epsilon\).}
Figure~\ref{fig:score-threshold} illustrates the score distribution during the local-scoring phase on the GSM8K dataset, where we observe a normal distribution of scores. 
The model seldom assigns very low scores (0-2 points). 
A lower score threshold \(\epsilon\) leads to an excessive number of reasoning chains proceeding to global evaluation; for instance, setting \(\epsilon\) to 3 results in over 95\% of reasoning chains moving to global evaluation. 
Conversely, a higher \(\epsilon\) enforces stricter filtering; setting \(\epsilon\) to 8 results in fewer than 10\% of reasoning chains moving forward to global evaluation, leading to many samples having only one reasoning chain in the global-evaluation phase. 
Some samples might even finish dynamic sampling without any reasoning chains proceeding to global evaluation. Therefore, we determine the score threshold \(\epsilon\) to be 6, which ensures a balance by maintaining high-quality reasoning chains and allowing a sufficient number to undergo global evaluation.

\vspace{-.5em}
\paragraph{Analysis of Termination Threshold \(\theta\).}
Figure~\ref{fig:termination-threshold} demonstrates the impact of various termination thresholds (\(\theta\)) on accuracy and computational cost. 
A threshold of \(\theta = 0\) implies that we select the answer associated with the highest-scoring reasoning chain as the final answer without sampling additional reasoning chains. 
While this approach incurs lower costs, it results in the poorest performance on both the GSM8K and AQuA datasets. 
This suggests that relying solely on the model's confidence in the highest-scoring reasoning chain does not guarantee its correctness. 
As the threshold increases, we observe a gradual improvement in accuracy. 
This indicates that imposing additional constraints and introducing new reasoning chains when necessary can aid the model in selecting the correct reasoning process. 
However, performance tends to saturate beyond a threshold of 2. 
We note that at a threshold of 4, more than 15\% of samples in the GSM8K dataset fail to produce a final answer even upon reaching the maximum number of sampled reasoning chains. 
Furthermore, excessively high thresholds also lead to significant increases in computational costs. 
Therefore, we establish the termination threshold \(\theta\) at 2, achieving an optimal balance between the accuracy of the outputs and the sampling costs of reasoning chains.

\vspace{-.5em}
\paragraph{Analysis of Batch Size \(b\).}
Figure~\ref{fig:batch-size} illustrates the impact of varying batch sizes (\(b\)) on accuracy and computational costs. 
During our analysis, samples exceeding the context window are excluded. 
We observe consistent performance improvements on both the GSM8K and AQuA datasets when evaluating multiple samples simultaneously, as opposed to assessing each sample individually. 
One possible explanation is that evaluating samples together allows the LLM to compare differences across reasoning chains, thereby providing more reliable scores. 
As the batch size increases, accuracy improves gradually until it reaches a batch size of 6, beyond which accuracy begins to fluctuate and even decline. 
At this point, the model's output becomes unstable, with some samples exceeding the model's context window, resulting in failed evaluations. 
Concurrently, we noted a gradual decrease in computational costs with increasing batch size, attributed to the reduced overhead of repetitive prompts. 
However, this trend starts to slow down when \(b > 2\). 
Therefore, we selected a batch size of \(b = 5\), which not only achieves optimal accuracy and lower computational costs but also avoids evaluation failures due to samples exceeding the model's context window.

\end{document}